\documentclass[10pt,journal,compsoc]{IEEEtran}
\ifCLASSOPTIONcompsoc
  \usepackage{cite}
\else
  \usepackage{cite}
\fi
\usepackage{graphicx}
\ifCLASSINFOpdf
\else
\fi
\usepackage{amsmath}
\ifCLASSOPTIONcompsoc
  \usepackage[caption=false,font=footnotesize,labelfont=sf,textfont=sf]{subfig}
\else
  \usepackage[caption=false,font=footnotesize]{subfig}
\fi
\usepackage{url}

\usepackage{xcolor}

\usepackage{amssymb}
\usepackage{mathtools}
\usepackage{enumitem}
\usepackage{booktabs}
\usepackage{multirow}
\usepackage{algorithm} 
\usepackage{algpseudocode} 
\usepackage{pifont}
\usepackage{float}
\newcommand{\cmark}{\ding{51}}
\newcommand{\xmark}{\ding{55}}
\newcommand{\algrule}[1][.2pt]{\par\vskip.5\baselineskip\hrule height #1\par\vskip.5\baselineskip}

\newcommand{\xhdr}[1]{\vspace{2pt} \noindent {\textbf{#1}}}

\begin{document}

%
\title{DROID: Driver-centric Risk Object IDentification}
%
%
%
%
\author{Chengxi~Li,
        Stanley H. Chan
        and~Yi-Ting~Chen
\IEEEcompsocitemizethanks{
\IEEEcompsocthanksitem C. Li is with Meta Platforms, Inc., Menlo Park, CA, 94025. Email: chengxili628@meta.com. 
\IEEEcompsocthanksitem S. H. Chan is with the Department
of Electrical and Computer Engineering, Purdue University, West Lafayette,
IN, 47906. Email: stanchan@purdue.edu
\IEEEcompsocthanksitem Y.-T. Chen is with National Yang Ming Chiao Tung University, Hsinchu, Taiwan. Email: ychen@nycu.edu.tw. 

\IEEEcompsocthanksitem A part of the work was done when C. Li was an intern and Y.-T. Chen was a research scientist at Honda Research Institute USA, San Jose, CA, USA. \protect}}
\IEEEtitleabstractindextext{
\begin{abstract}
Identification of high-risk driving situations is generally approached through collision risk estimation or accident pattern recognition. In this work, we approach the problem from the perspective of subjective risk. We operationalize subjective risk assessment by predicting driver behavior changes and identifying the cause of changes. To this end, we introduce a new task called driver-centric risk object identification (DROID), which uses egocentric video to identify object(s) influencing a driver’s behavior, given only the driver’s response as the supervision signal. We formulate the task as a cause-effect problem and present a novel two-stage DROID framework, taking inspiration from models of situation awareness and causal inference. A subset of data constructed from the Honda Research Institute Driving Dataset (HDD) is used to evaluate DROID. We demonstrate state-of-the-art DROID performance, even compared with strong baseline models using this dataset.
Additionally, we conduct extensive ablative studies to justify our design choices.
Moreover, we demonstrate the applicability of DROID for risk assessment.
\end{abstract}
}

\maketitle

\IEEEdisplaynontitleabstractindextext

%
\IEEEpeerreviewmaketitle

\IEEEraisesectionheading{\section{Introduction}\label{sec:introduction}}

\IEEEPARstart{M}{o}re than 1.3 million people die in road accidents worldwide every year, or approximately 3\,700 people per day~\cite{who}. Road traffic accidents are among the leading causes of non-natural death around the world. The majority of these accidents are due to driver errors, such as exercising poor awareness~\cite{nhtsa}. To reduce the number of accidents, developing intelligent driving systems such as advanced driver assist systems (ADAS) that identify high-risk situations is in urgent need.

\begin{figure}[t!]
\includegraphics[width = \columnwidth]{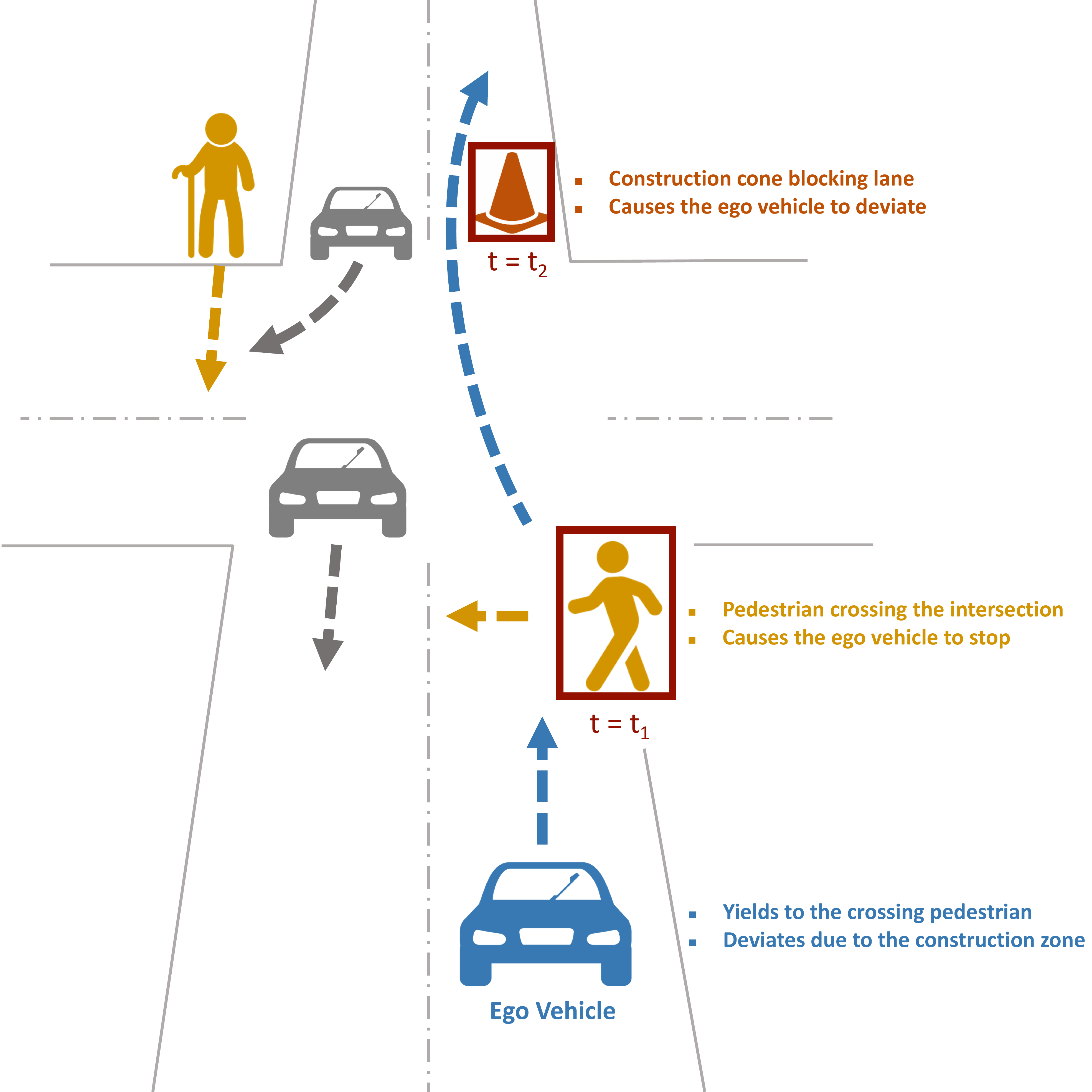}
\vspace{-18pt}
\caption{Human drivers perceive scenes, assess risks, make a plan, and take actions in different traffic situations.
Risk assessment, identifying hazards and risk factors that have the potential to cause harm, is indispensable for driving safety.
{In this work, we focus on the prediction of driver behavior changes and identification of the cause of
changes. 
We cast the problem as a cause-effect problem.
We propose a novel computational framework that identifies traffic participants (cause, such as the crossing pedestrian shown) making drivers react (effect).} 
}
\label{fig:traffic_example}
\end{figure}

{This problem of risk assessment has been studied extensively in the literature~\cite{Minderhoudttc2001,Althoff_srs_iv2008,worrall2010improving,althoff2011comparison,greene2011efficient,Lefevre_iros2012,lawitzky2013interactive,risk_assessment_Lefevre_ROBOMECH,hariyono2016estimation,Roth_path_iv2016,Paigwar_probablityrisk_iv2020,chinea2007risk,salim2007collision,chan2016anticipating,suzuki2018anticipating,yao_unsupervised_iros2019,You_eccv2020,Herzig_iccvw2019,DADA_2021,riskassessment}, and has been approached by modeling risk as objective or subjective.
In this work, we focus on subjective risk~\cite{subjectiverisk}, i.e., the driver’s own perceived risk, which is an output of the driver’s cognitive process.
For example, in Fig.~\ref{fig:traffic_example},  the driver reacts to the crossing pedestrian (i.e., the driver slows down) when passing through the intersection. The driver then reacts to the construction cone (i.e., deviates into the oncoming lane). In these daily tasks, drivers constantly observe traffic situations and plan accordingly to avoid potential hazards. 
Understanding driver behavior, i.e., when and why a driver reacts to a situation, is critical to the development of intelligent driving systems.
We propose to operationalize subjective risk assessment as the prediction of changes in driver behavior and identification of the cause of such changes. We then discuss the computational framework corresponding to this operationalization.} 

\begin{figure}[t!]
\includegraphics[width = \columnwidth, page=2]{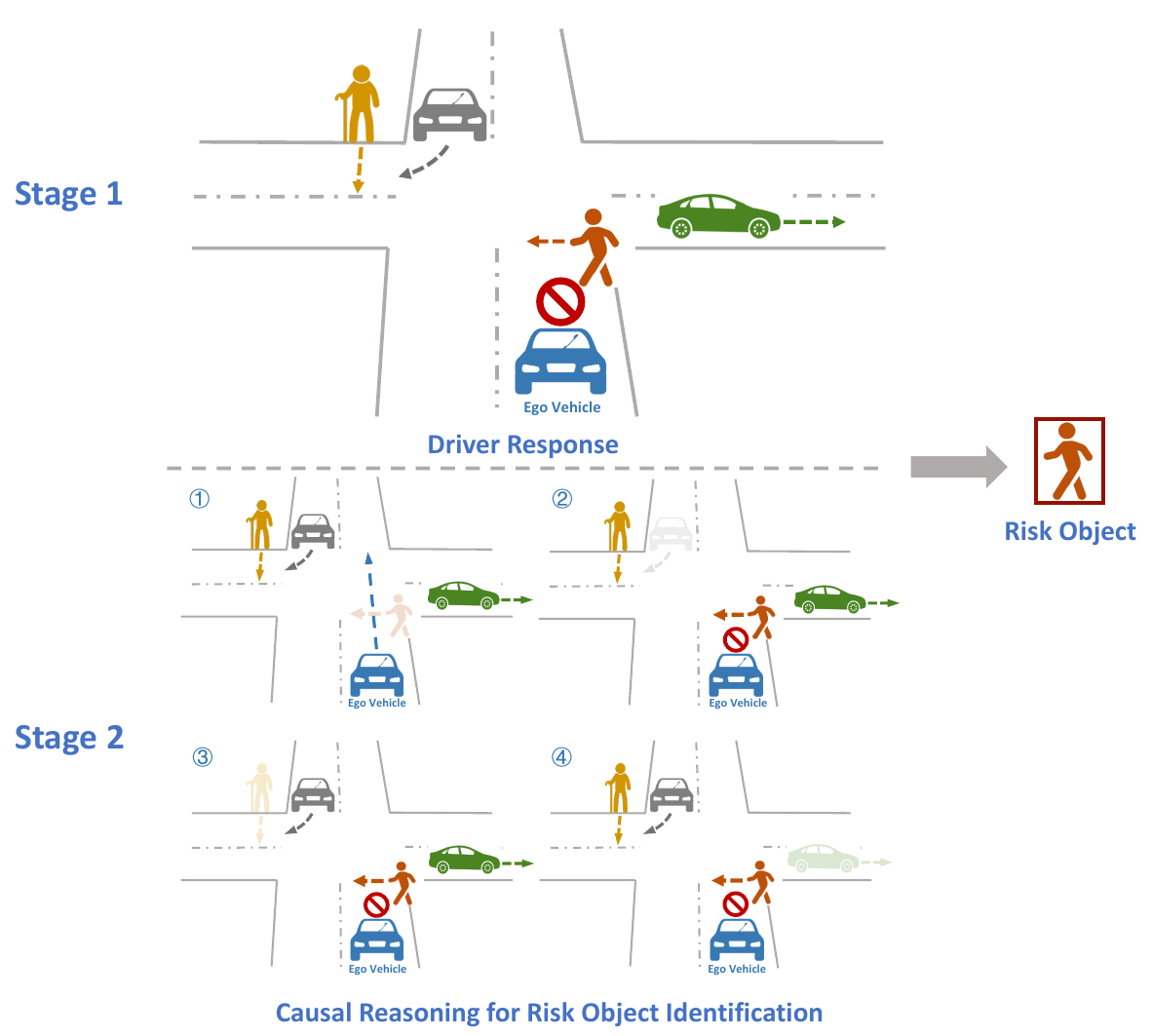}
\vspace{-18pt}
\caption{A conceptual diagram of the proposed two-stage DROID framework. We first predict driver response in a given situation.
To identify object(s) influencing driver behavior, we intervene in the input observation by removing a traffic participant at a time (i.e., simulating a situation without the traffic participant), and predict the corresponding driver response.
For instance, removing the crossing pedestrian changes driver response (effect).
The effects of removing other traffic participants remain the same.
We conclude that the crossing pedestrian is the risk object (cause).}
\label{fig:illustration}
\end{figure}

A natural question arises: \textit{What changes drivers' behavior?}
{We propose a new task called \textit{driver-centric risk object identification} (DROID), which uses egocentric video collected from front-facing cameras to identify object(s) influencing a driver’s behavior, given only the driver’s response as the supervision signal.
Note that risk object identification from egocentric videos is crucial for safety systems such as ADAS, where front-facing cameras are the primary device.
The proposed setting is more challenging than existing tasks because the latter utilize either human annotations of object importance~\cite{Spain_importance_eccv2008,Ohn-Bar_object_importance_2017,Gao_goal_icra2019,Zhang_graph_important_icra2020}, risk regions~\cite{Zeng_risk_cvpr2017,suzuki2018anticipating,Herzig_iccvw2019}, or human gaze patterns~\cite{Alletto_Dreye_cvprw2016,Xia_ACCV_2018,Xia_attention_wacv2020,Pal_semantic-gaze_cvpr2021,Baee_MEDIRL_iccv2021} as supervision signals for identification.
In contrast, the proposed task must identify driving-related risk or important objects from the driver's response.
}

{To address this challenging task, we formulate the problem as cause-effect \cite{Pearl_causality_2009}, and propose a novel two-stage framework.
The core concept of the framework is depicted in Fig.~\ref{fig:illustration}.
In the first stage, a driver model learns to predict driver response in a given situation.
To distinguish whether the driver’s behavior is influenced, we cast the problem as a binary classification, predicting the response of drivers to be \textit{Stop} or \textit{Go}.
Future works could further refine the granularity of driver response {}--{} for example, by including driver actions, such as stepping lightly or heavily on the brake pedal \cite{Kokubun_assessment_2005}.}
To accurately predict the response of a driver, a
driver behavior model should capture complicated spatio-temporal interactions between a driver and traffic elements (e.g., vehicles, pedestrian, and lane markings).
We propose a novel {driver behavior model} motivated by the model of situation awareness (SA)~\cite{Endsley_SA_2000}.
Specifically, the proposed model predicts driver responses by encapsulating the driver's goal (i.e., driver intention), perception (i.e., elements of the environment), comprehension (i.e., interactions between driver and \textit{Thing} objects and interactions between driver and \textit{Stuff} objects in 3D), and projection (i.e., intention-aware interaction forecasting).
\textit{Thing} and \textit{Stuff} objects are defined in Sec.\ref{subsec:perception}.

In the second stage of the framework, given a \textit{Stop} response (i.e., driver behavior is influenced by certain objects), we intervene in input video by removing a tracklet at a time and inpainting the removed area in each frame to simulate a scenario without the presence of the tracklet. 
The same {driver behavior model} is applied to predict the effect of the removal.
The process iterates through all tracklets and records the corresponding effects on driver response. 
Note that we assume that the cause of driver response change is either vehicles or pedestrians.
The tracklet whose removal causes a maximum response change is the risk object.

Our work differs from existing works threefold:
\begin{enumerate}
\item {\textbf{Operationalization of Subjective Risk Assessment:} We propose to operationalize subjective risk assessment by predicting driver behavior changes and identifying the cause of such changes.}
\item \textbf{Task Formulation:} A new task called {DROID} is introduced, which aims to identify object(s) influencing driver’s behavior from egocentric videos, given only the driver’s response as the supervision signal. 
\item \textbf{Methodology:} A causal inference-based framework is proposed to identify risk objects.
\end{enumerate}


In this work, we make the following substantial extensions to our early results\cite{li2020make}:
\begin{enumerate}
\item We re-design the {driver behavior model} substantially to predict driver response by modeling driver decision processes via the model of SA~\cite{Endsley_SA_2000}.
\item We systematically benchmark three different tasks on the constructed dataset, i.e., driver response prediction, driver intention prediction, and DROID.
\item We conduct thorough ablative studies to justify the architectural designs.
\end{enumerate}

\section{Related Work}

{\xhdr{Vision-based Risk Assessment.}
Numerous attempts have been made to build better systems that can robustly assess high-risk situations to reduce traffic fatalities. 
Existing objective risk assessment algorithms aim to estimate collision risks by computing of time-to-X (e.g., time-to-collision) ~\cite{Hayward1972NEARMISSDT,Minderhoudttc2001,worrall2010improving};
%
%
predicting traffic participants' future trajectories~\cite{bi2019joint,Salzmann_trajecron_eccv2020,cui2020deep,kawasaki2020multimodal,huang2020diversitygan,mangalam2020not,liu2020spatiotemporal,EvolveGraph,yau2021graph} for  collision checking~\cite{Althoff_srs_iv2008,althoff2011comparison,greene2011efficient,hariyono2016estimation,Roth_path_iv2016,Paigwar_probablityrisk_iv2020}; or detecting and anticipating traffic accidents~\cite{chan2016anticipating,suzuki2018anticipating,yao_unsupervised_iros2019,You_eccv2020,Herzig_iccvw2019,DADA_2021}.
%
%
Reliable state estimation from images of traffic participants and environments are the prerequisite for vision-based collision risk estimation-based methods.
Significant advances in state estimation algorithms such as object detection and tracking~\cite{HeCVPR2017,ETDR2020,yolov4,Wojke2017simple,Wojke2018deep,Yang_VIS2019}, depth estimation~\cite{Zhou_cvpr2017,Guizilini,Jung_iccv2021}, road topology modeling~\cite{Wang_top_CVPR2019,Philion_lift_eccv2020,Tian_roadgraph_icra_2021}, and trajectory prediction~\cite{TraPHic_CVPR2019,titan_CVPR2020,Neumann_CVPR2021} are observed in the literature.
The authors of~\cite{Janai2020} provide a comprehensive review of vision-based algorithms for traffic scene understanding.}
%

{In this work, instead of estimating collision risks, we focus on subjective risk assessment~\cite{subjectiverisk}, an output of the driver’s cognitive process.
Existing works~\cite{8370754,Yurtsever_itsc2019,a8917457,riskassessment} formulate subjective risk assessment as risk-level classification, which requires human annotators to label the risk level of a scenario (e.g., lane change).
This work looks into a different formulation for subjective risk assessment.
Specifically, we utilize driver behavior changes as direct outputs of driver’s cognitive process.
The formulation mitigates the subjectivity found in the existing strategy.  
Furthermore, we also identify the cause of such changes, which are not considered in~\cite{8370754,Yurtsever_itsc2019,a8917457,riskassessment} and are essential for planning and decision making.
To this end, we propose a new task called DROID, which aims to identify object(s) influencing driver’s behavior from egocentric videos, given only the driver’s response as the supervision signal.
We propose an end-to-end trainable framework that leverages the SA model~\cite{Endsley_SA_2000} to encapsulate state estimation, situation understanding, and future prediction for driver response prediction.
Furthermore, we incorporate causal inference to improve the performance of prediction and identify risk objects.
We empirically demonstrate the effectiveness of our proposed method for DROID. 
We hope that our findings will pave the way for a tight integration of causal reasoning and vision-based risk assessment, a largely under-explored yet critical path towards reliable intelligent driving systems. 
}

{\xhdr{Vision-based Driver Behavior Modeling.}
Driver behavior modeling has been studied extensively in the intelligent vehicle community.
Driver behavior is inherently multimodal and difficult to predict, but effective modeling of driver behavior is indispensable to enabling safe and robust intelligent driving systems (e.g., advanced driver assist systems).
Detailed reviews of existing driving behavior modeling can be found~\cite{Michon1985,Doshi_tactical_driver,Eshed_lookinghuman2016}.}
{While significant advances have been shown, vision-based driver behavior models that use egocentric video to predict driver's intention and response have not been widely explored.
We review vision-based end-to-end driving models and interaction modeling from egocentric view prediction that are relevant to the proposed driver behavior model in the following.} 

{End-to-end driving models have drawn a considerable attention in the vision community~\cite{AlVINN_1989,Bojarski_nvidia_2016,codevilla2018end,Codevilla_iccv2019,Chen_corl2019,Zhou_depth_action_2019,Ohn-Bar_situation_cvpr2020,Behl_iros2020,Prakash_aggregation_cvpr2020}. 
Recent driving models~\cite{Zhou_depth_action_2019, Behl_iros2020} have shown the effectiveness of incorporating semantics and depth as intermediate representations for driving.
The proposed driver behavior model considers both cues as well.
Specifically, we utilize semantics to differentiate different traffic participants (i.e., \textit{Thing} and \textit{Stuff}, defined in Sec.\ref{subsec:perception}). In addition, depth is used to calculate spatial distances between ego vehicle and traffic participants for relational modeling using graph convolution networks (GCNs)~\cite{Kipf_GCN_iclr2017}.   
The conditional imitation learning framework~\cite{codevilla2018end,Sauer2018CORL,codevilla2019exploring} is introduced to condition imitation learning on a high-level command input. Specifically, the framework enables a trained driving policy that can respond to a navigational command.
Motivated by this, our proposed driver behavior model conditions driver response prediction on driver intention.
We empirically show that the proposed designs are effective for driver response prediction. Moreover, we demonstrate the trained driver behavior model is valuable for risk object identification.
Note that the proposed driver behavior model is complementary to end-to-end driving models.
Novel architectural designs of end-to-end driving modeling may benefit vision-based driver behavior modeling and vice versa.}

\begin{figure*}[h!]
\includegraphics[width=\textwidth]{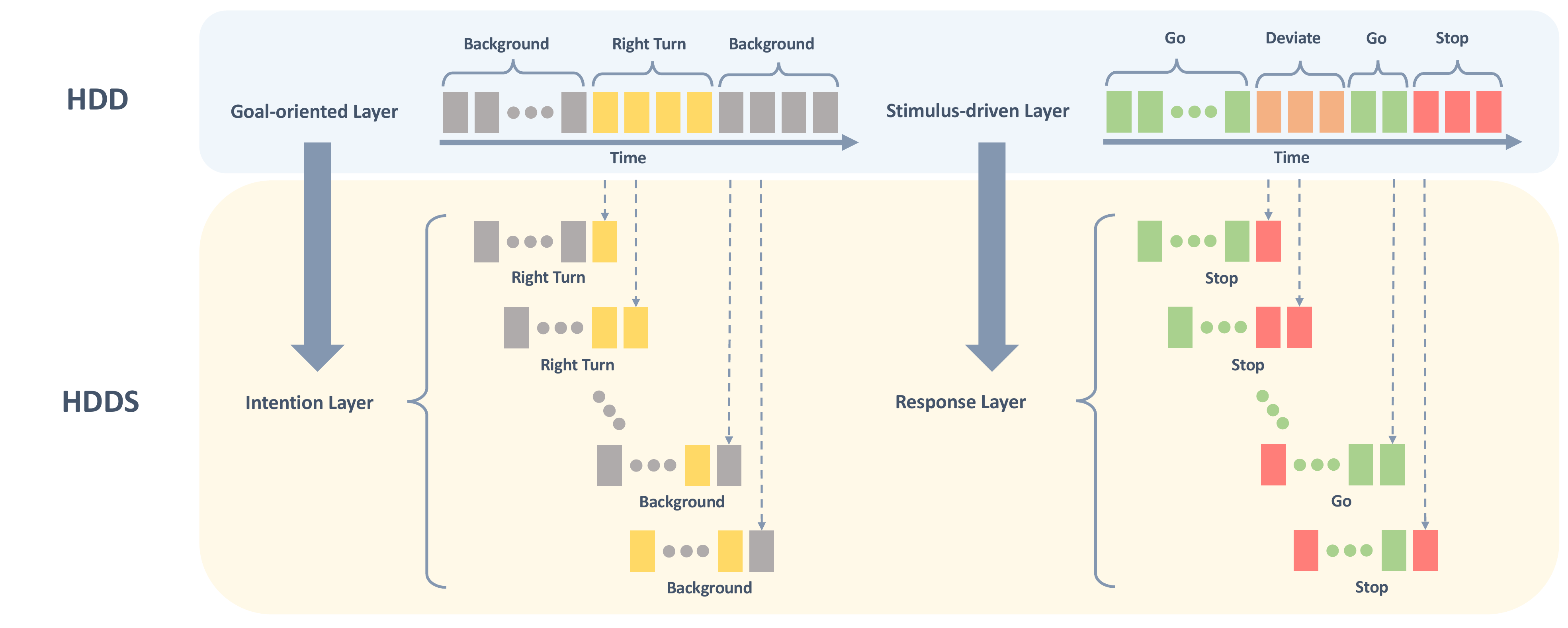}
    \caption{Our data subset (HDDS) for DROID. 
    We construct the subset from the Honda Research Institute Driving Dataset (HDD).
    In particular, we introduce two layers, i.e., \textbf{Driver Intention} and \textbf{Driver Response}, in HDDS.
    Further detail of the two layers can be found in Sec.\ref{sec:roi_dataset}.
    To obtain \textbf{Intention} labels, we form $n$-frame clips, and the corresponding \textbf{Intention} label of each clip is the last frame's label defined in the \textbf{Goal-oriented} layer. 
    A similar procedure is applied to construct the \textbf{Response} layer (as shown on the right-hand side of the figure). 
    Notice that both \textit{Stop} and \textit{Deviate} annotated in HDD are merged into \textit{Stop} in our dataset.
    A \textbf{Cause} layer (not shown) identifies four possible reasons for \textit{Stop} or \textit{Deviate}.
    }
    \label{fig:hdd_roi}
\end{figure*}

{Interaction modeling from an egocentric view has been explored in pedestrian behavior prediction~\cite{titan_CVPR2020,liu2020spatiotemporal}; driver attention prediction~\cite{DADA_2021,Baee_MEDIRL_iccv2021}; driver intention prediction~\cite{Jain_driver_maneuver_iccv2015,Jain_rnn_icra2016,Jain_structurernn_cvpr2016}; important object detection~\cite{Zhang_graph_important_icra2020}; and human-human or human-object interaction~\cite{Ma_interaction_cvpr2018,Li_dual_cvpr2019,Li_realtion_pami2022}.}
{Our work is closely related to~\cite{Jain_structurernn_cvpr2016,Li_dual_cvpr2019,Zhang_graph_important_icra2020,Li_realtion_pami2022} because these works explicitly model interactions between ego (e.g., camera wearer~\cite{Li_realtion_pami2022}) and interactors (e.g., objects in kitchen~\cite{Damen2018EPICKITCHENS}).
The explicit modeling between ego and interactors is important for understanding the behavior of the camera wearer or ego vehicles~\cite{Jain_structurernn_cvpr2016,Li_dual_cvpr2019}.
In the Brain4Car project~\cite{Jain_structurernn_cvpr2016}, Jain et al. propose to combine spatio-temporal graphs and Long
Short-term Memory (LSTM) to capture the relationship between the driver and inside/outside context for driver intention prediction.
In~\cite{Li_dual_cvpr2019,Li_realtion_pami2022}, the authors exploit LSTM~\cite{hochreiter1997long} to model the relation between ego and interactors.
Instead of using LSTM, Zhang et al.~\cite{Zhang_graph_important_icra2020} leverage GCNs~\cite{Kipf_GCN_iclr2017} to model interactions between ego vehicle and traffic participants. 
In this work, we use GCNs to model interactions. This work differs from those previously described in that we introduce two types of interactions (i.e., interactions between ego vehicle and \textit{Thing} objects, and interactions between driver and \textit{Stuff} objects in 3D) for driver response prediction and risk object identification.
Empirically, we demonstrate the importance of the proposed architecture for the targeted applications.
} 

\xhdr{Causality in Computer Vision.}
Computer vision research has proliferated over the past decades due to the advance in deep learning algorithms.
However, current deep learning models may suffer from spurious correlation~\cite{pearl2016causal} as a result of ignoring causality in data, although in fact, humans perceive the causality of the physical world.
To address the issue, recent studies\cite{NairInduction2019, LiCausalDiscovery2020,QiCausalCVPR2020,WangCommonCVPR2020,YangDeconfound2020} explicitly consider the concept of causality in deep learning architectural designs.

\begin{table*}[t]
    \small
    \centering
    \resizebox{\textwidth}{!}{
        \begin{tabular}
            {l c c c c c c c c c c c c c c c}
            \toprule
             \multirow{3}{*}{ \begin{tabular}{@{}c@{}} \\Split \end{tabular}} & 
              &
             \multicolumn{12}{c}{Intention}  & 
             \multicolumn{2}{c}{Response}  \\
             \cmidrule(lr){3-14} \cmidrule(lr){15-16}
             &
             &
             BG & 
             IP& 
             LT & 
             RT & 
             LLC & 
             RLC & 
             LLB & 
             RLB & 
             CP &
             RP & 
             MG &
             UT & 
             STP& 
             G \\
             \midrule
             Train  & & 737\,949  & 48\,933  & 21\,819 & 19\,824 & 4\,815 & 4\,386 & 1\,833 & 717 & 2\,364 & 588 & 1\,182 & 2\,001 & 184\,890 & 661\,521 \\
            \midrule
            Test1   & & 236\,622 & 17\,772 & 7\,017 &  6\,195 & 1\,098 & 1\,212 & 435 & 324 & 432 & 123 &327 & 432 & 63\,314 & 208\,675      \\        
            \midrule
             & Cause & \\
            \cmidrule(lr){2-2}
            
            \multirow{4}{*}{ \begin{tabular}{@{}c@{}} \\Test2 \end{tabular}}& Congestion & 98 & / & / & / & 1 & / & / & / & / & / & / & / & 99 & / \\
            & Crossing Pedestrian & 62& 15 & 5 & / & / & / & / & / & 2 & / & / & / & 84 & / \\
            & Crossing Vehicle & 263 &  2 & 7 & 35 & / & / & / & 4 & / & / & / & / & 311 & / \\
            & Parked Vehicle & 120 & 3 & / & / & 9 & 4 & / & / & / & / & / & / & 136 & / \\
            \cmidrule(lr){2-16}
            & All & 543 &  20 & 12 & 35 & 10 & 4 & / & 4 & 2 & / & / & / & 630 & / \\ 
            \midrule
            \multicolumn{16}{l}{\multirow{3}{*}{\begin{tabular}{@{}l@{}}Intention: (BG) background, (IP) intersection passing, (LT) left turn, (RT) right turn,(LLC) left lane change, (RLC) right lane change, (LLB) left lane branch, \\ (RLB) right lane branch, (CP) crosswalk passing, (RP) railroad passing, (MG) merge, (UT) u-turn.\\Response: (STP) stop, (G) go. \end{tabular}}}\\
            \\
            \\
            \bottomrule
        \end{tabular}
        }
\vspace{8pt}
\caption{Statistics (annotated frames) of HDDS.}\label{table:data_statistics}
\end{table*}

To our best knowledge, we are among the first to apply causal inference to egocentric images captured in driving scenes.
Kim et al.~\cite{kim2017interpretable} propose a causality test as a means to  identify  input regions influencing the output their driving model.
We also employ causal inference similar to the causality test of Kim et al.\cite{kim2017interpretable}. 
However, the purpose of causal inference in this work is to identify risk objects that cause drivers to change behaviors.
Moreover, we design a simple but effective data augmentation strategy using causal intervention.
This leads to a more robust driving model.

{Haan et al.~\cite{Haan_causalconfusion_nips2019} propose to incorporate functional causal models~\cite{Pearl_causality_2009} into imitation learning to address the issue of ``causal misidentification," a phenomenon in which accessing to more information leads to unsatisfactory generalization performance in the presence of distributional shift
in imitation learning.}
%
%
{Samsami et al.\cite{samsami2021causal} extend~\cite{Haan_causalconfusion_nips2019} to address inertia and collisions found in imitation learning-based autonomous driving policy. 
A causal approach that mitigates the impact of distribution shifts in motion forecasting is studied in a recent work~\cite{Liu_causaltrajectory_cvpr2022}.}

{Our work is complementary to~\cite{Haan_causalconfusion_nips2019,samsami2021causal,Liu_causaltrajectory_cvpr2022}.
Specifically, the focus of these models~\cite{Haan_causalconfusion_nips2019,samsami2021causal,Liu_causaltrajectory_cvpr2022} is to improve the robustness under distribution shifts via causal inference.
In this work, causal inference is used to synthesize counterfactual scenarios to improve the performance of driver behavior prediction and identify risk objects.
}
%
%

\section{Dataset}
\label{sec:roi_dataset}
To study DROID, a dataset with diverse reactive scenarios (i.e., drivers react to potential hazards while navigating to their goals) is indispensable. 
For instance, when human drivers intend to turn left at an unprotected intersection, they react (e.g., slowing down or stopping) to certain traffic participants to avoid dangerous situations.
{Existing datasets \cite{chan2016anticipating,suzuki2018anticipating,yao_unsupervised_iros2019,You_eccv2020} are used to study traffic accidents. However, these datasets cannot be utilized to study DROID, which aims to discover the causal relationship between risk object and driver behavior change, because they only have accident data. Thus, we leverage the Honda Research Institute Driving Dataset (HDD)  \cite{RamanishkaCVPR2018}\footnote{The dataset is available at \url{https://usa.honda-ri.com/HDD}} to construct a subset of data for DROID.}

Fig.~\ref{fig:hdd_roi} illustrates how we construct the subset from the HDD dataset, hereafter referred to as HDD Subset (HDDS).
The \textbf{Goal-oriented} layer defined in the HDD dataset denotes tactical driver behavior such as \textit{right turn}, \textit{left turn}, or \textit{lane change}.
As shown in Fig.~\ref{fig:hdd_roi}, each frame is labeled with either a goal-oriented or background action.
To obtain the \textbf{Intention} of an $n$-frame clip (the parameter $n$ is 20 in our implementation), we use the last frame's label of the \textbf{Goal-oriented} layer as the \textbf{Intention} label.
While performing a tactical behavior, drivers might have to \textit{stop} or \textit{deviate} due to traffic participants or obstacles. 
We extend the \textbf{Stimulus-driven} actions, i.e., \textit{Stop} and \textit{Deviate}, defined in the HDD dataset, as the \textbf{Response} label.
Note that both \textit{Stop} and \textit{Deviate} are merged into \textit{Stop}, as depicted in Fig.~\ref{fig:hdd_roi}. 
The rest of the frames are labeled as \textit{Go}. 
The HDD dataset also annotates a \textbf{Cause} layer to explain the reason for \textit{Stop} and \textit{Deviate} actions. 
We obtain the \textbf{Test2} set of the HDDS by selecting frames from the four \textbf{Cause} scenarios, i.e., \textit{Congestion}, \textit{Crossing Pedestrian}, \textit{Crossing Vehicle} and \textit{Parked Vehicle}.
{Moreover, in the \textbf{Test2} set, we use annotated bounding boxes of risk objects (i.e., object[s] influencing driver's behavior) from HDD for DROID benchmarks. In \cite{RamanishkaCVPR2018}, given an interactive scenario (i.e. \textbf{Cause} layer is labeled as \textit{Crossing Vehicle}), Ramanishka et al. work with a third-party annotation company to have two human annotators label the closest object with the same category (vehicle) labeled in the \textbf{Cause} layer. Then, there is a third person who checks the consistency between the two annotators.}
For our DROID benchmarks, we focus on scenarios in which drivers react to vehicles or pedestrians.

HDDS has 184\,890 frames for training driver response and intention predictors.
Two test sets are constructed for driver response prediction and DROID, respectively.
The \textbf{Test1} split has 63\,314 frames for both driver response and intention benchmarks. 
The \textbf{Test2} has 630 frames (i.e., 630 different risk objects) covering four reactive scenarios, i.e., \textit{Congestion}, \textit{Crossing Pedestrian}, \textit{Crossing Vehicle}, and \textit{Parked Vehicle} for DROID benchmarks. 
Detailed statistics are shown in Table~\ref{table:data_statistics}.
{Note that in daily driving, drivers react to diverse traffic participants, e.g., bicyclists, motorcyclists, jaywalkers, traffic signs, construction areas, and so on. A new dataset that covers these interaction types would be invaluable for benchmarking DROID. We leave it for future work.}

\section{Problem Formulation}\label{section:problem_formulation}
\begin{figure*}[t]
\includegraphics[width=\textwidth,page=1]{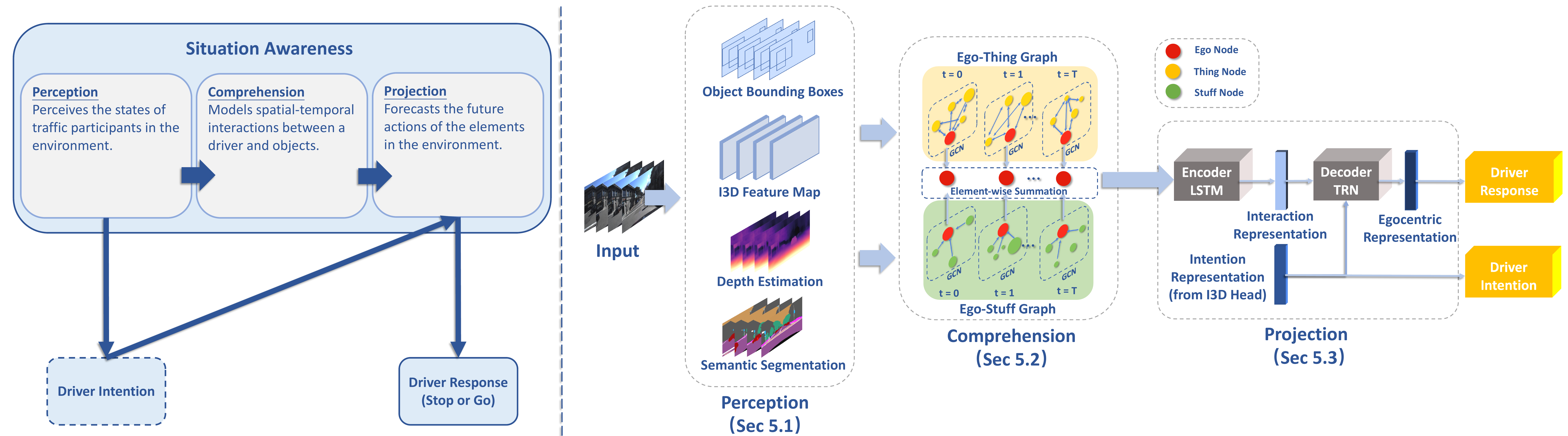}
    \caption{An overview of the proposed driver behavior model for driver response prediction (right). 
    The proposed architecture is motivated by the model of situation awareness~\cite{Endsley_SA_2000} (left).
    Given a video clip, 3D convolutions (I3D \cite{CarreiraCVPR2017}), object detection, semantic segmentation, and depth estimation are applied to obtain states of traffic participants in a traffic environment at the \textit{Perception} stage (Sec.~\ref{subsec:perception}). 
    At the \textit{Comprehension} stage, an \textit{Ego-Thing Graph} and an \textit{Ego-Stuff Graph} are constructed to model spatiotemporal interactions between a driver and traffic participants (Sec.~\ref{subsec:comprehension}).
    In this work, we categorize traffic participants into two types, i.e., \textit{Thing} and \textit{Stuff}.
    The details are discussed in Sec.~\ref{subsec:comprehension}.
    The final stage, \textit{Projection} (Sec.~\ref{subsec:projection}), forecasts future interactions between driver and traffic participants for driver response prediction. 
    Frame-wise interactions obtained from \textit{Ego-Thing Graph} and \textit{Ego-Stuff Graph} are fused and fed into an encoder LSTM to form interaction representation. 
    Intention representation obtained from the I3D head and interaction representation are sent to a decoder Temporal Recurrent Network, or TRN (the architecture is shown in Fig.~\ref{fig:trn}) to predict driver response. 
    %
    }
    \label{fig:driving_model}
\end{figure*}

Given a reactive scenario with $T$ RGB images $I \coloneqq \{I_1, I_2,\cdots, I_T\}$, the goal is to identify the object influencing driver response in the last frame (DROID task).

We formulate this task as a cause-effect problem~\cite{Pearl_causality_2009}.
Specifically, a two-stage framework is proposed to identify the cause (i.e., the risk object) of an effect (i.e., driver behavior change) via the \textit{Situation Awareness-based Driver Behavior Model} and \textit{Causal Reasoning for DROID}.
We discuss the methodology in the following.

\section{Driver Behavior Model}
\label{sec:driving_model}
An overview of the proposed driver behavior model is depicted in Fig.~\ref{fig:driving_model}.
To predict the response of a driver, a driver behavior model should capture complicated spatio-temporal interactions between a driver and traffic participants.
We propose a novel driver behavior model motivated by the model of SA~\cite{Endsley_SA_2000}.
Fig.\ref{fig:driving_model} (left) provides an overview of the SA model.
Specifically, the proposed model encapsulates the four essential components defined in SA for driver response prediction: goal/objective (i.e., driver intention), perception (i.e, elements of a traffic environment), comprehension (i.e., interactions between driver and \textit{Thing} objects, and interactions between driver and \textit{Stuff} objects in 3D), and projection (i.e., intention-aware interaction forecasting).
The detail of each component is discussed in the following.

\subsection{Perception}
\label{subsec:perception}
Perception plays an essential role in the SA model~\cite{Endsley_SA_2000}.
This component incorporates the status, attributes, and dynamics of relevant traffic participants of a traffic environment.
Specifically, given $T$ RGB images, we apply object detection~\cite{HeCVPR2017} and semantic segmentation~\cite{rotabulo2017place} to obtain \textit{Thing} and \textit{Stuff} objects, respectively.
{Previous work \cite{forsyth1996finding,heitz2008learning,caesar2018coco} has different definitions of \textit{Thing} and \textit{Stuff} based on the spatial extent or shape of objects.}
In this work, we distinguish \textit{Stuff} objects from \textit{Thing} objects by evaluating whether states of an object can be influenced by other objects.
If yes, we categorize the object as a \textit{Thing} object.
{
For instance, cars are \textit{Thing} objects since they stop or yield to obstacles.
A traffic light turns red or green by itself, so is categorized as a \textit{Stuff} object.}
In addition to detection and segmentation, we perform object tracking using Deep SORT \cite{Wojke2017simple} and depth estimation~\cite{Ranftl2022}.

\subsection{Comprehension}
\label{subsec:comprehension}
We interpret \textit{Comprehension} as the spatio-temporal interactions between the driver and \textit{Thing} objects, and interactions between the driver and \textit{Stuff} objects in the 3D world.
Note that a thorough modeling of \textit{Comprehension} is beyond the scope of this work.
Specifically, we construct two graphs, i.e., \textit{Ego-Thing Graphs} and \textit{Ego-Stuff Graphs}, modeled with GCNs~\cite{Kipf_GCN_iclr2017}.
The details of each graph are discussed below.
%
%
{Note that we choose the interaction modeling proposed in~\cite{Chengxi2020} because the model explicitly model interactions among drivers, traffic participants, and road scene infrastructure from egocentric images.} 
In this work, we extend the modeling for driver response prediction.

\subsubsection{Ego-Thing Graph}
\label{subsection:ego-thing graph}

The \textit{Ego-Thing Graph} is designed to model interactions between a driver and \textit{Thing} objects.

\noindent{\textbf{Graph Definition.}} 
We denote a sequence of frame-wise \textit{Ego-Thing Graphs}  as $\mathbf{G}^{ET} = \{\mathbf{G}^{ET}_t | t =1,\cdots,T\}$, where $T$ is the number of frames, and $\mathbf{G}^{ET}_{t}\in \mathbb{R}^{(K+1) \times (K+1) }$ is the \textit{Ego-Thing} affinity matrix at frame $t$ encoding pair-wise interactions among \textit{Thing} objects and \textit{Ego}.
Specifically, ${G}^{ET}_t(i,j)$ denotes the influence of object $j$ on object $i$.  
A node $i$ at time $t$ is represented by two types of features $(\mathbf{x}^{t}_i,\mathbf{p}^t_i)$, where $\mathbf{x}^{t}_i$ represents the appearance feature, and $\mathbf{p}^t_i \in \mathbb{R}^{1 \times 3}$ is the 3D location of the $i$-th object in respect to \textit{Ego} in a local frame. 

\noindent{\textbf{Node Feature Extraction.}} 
\textit{Thing} objects are \textit{car}, \textit{person}, \textit{bicycle}, \textit{motorcycle}, \textit{bus}, \textit{train}, and \textit{truck}. 
Given bounding boxes obtained from object detection \cite{HeCVPR2017}, we keep $K$ top-scoring detected boxes. 
The parameter K is set to 20  {empirically, as most frames in HDDS have no more than 20 objects}.
There are $K+1$ objects, where index $i=1,2,\cdots,K$ corresponds to \textit{Thing} objects, and index $K+1$ corresponds to \textit{Ego}.  
The appearance feature for $i$-th object is denoted as $\mathbf{x}^{t}_i \in \mathbb{R}^{1 \times D}, i = 1, 2, \cdots, K, K+1$.
RoIAlign \cite{HeCVPR2017} and max pooling are applied to obtain the appearance features of \textit{Thing} objects.
The appearance of \textit{Ego} is obtained by the same procedure as \textit{Thing} objects, but with a frame-size bounding box. 

\noindent{\textbf{Relational Modeling.}} 
We consider both appearance features and distance constraints, motivated by \cite{Wu_groupGCN_cvpr2019} in relational modeling. 
An edge $G_t^{ET}(i,j)$ is defined as:
\begin{equation}
    {G}^{ET}_t(i,j) = \frac{f_s(\mathbf{p}^t_i,\mathbf{p}^t_j)\text{exp}[f_a(\mathbf{x}^{t}_i,\mathbf{x}^{t}_j)]}{\sum_{j=1}^{K+1}  f_s(\mathbf{p}^t_i,\mathbf{p}^t_j)\text{exp}[f_a(\mathbf{x}^{t}_i,\mathbf{x}^{t}_j)]} \\,
    \label{eq:1}
\end{equation}
where $f_a(\mathbf{x}^{t}_i,\mathbf{x}^{t}_j)$ indicates an appearance relation, and $f_s(\mathbf{p}^t_i,\mathbf{p}^t_j)$ denotes relative distance between $i$-th and $j$-th object, respectively. 
The softmax function is used to normalize an affinity matrix $G_t^{ET}$.
An appearance relation is defined as below:
\begin{equation}
    f_a(\mathbf{x}^{t}_i,\mathbf{x}^{t}_j) = \frac{\phi(\mathbf{x}^{t}_i)^{\text{T}} \phi '(\mathbf{x}^{t}_j)}{\sqrt{D}} \\,
\end{equation}
where $\phi(\mathbf{x}^{t}_i) =\mathbf{w}\mathbf{x}^{t}_i $ and $\phi '(\mathbf{x}^{t}_j) = \mathbf{w}'\mathbf{x}^{t}_j$. 
Both $\mathbf{w} \in \mathbb{R}^{D \times D}$ and $\mathbf{w}'\in \mathbb{R}^{D \times D}$ are trainable parameters.
$\sqrt{D}$ is a normalization factor.

In addition to the appearance relation, we also consider spatial constraint via calculating a relative distance between a pair of objects.
Specifically, we unproject the center of a \textit{Thing} object's bounding box to 3D space~\cite{Ranftl2022}. For \textit{Ego}, we unproject the middle-bottom pixel of the frame to 3D space.
Given a 2D coordinate $(u_i^t,v_i^t)$ of $i$-th object at time $t$, the corresponding 3D coordinate $( x_i^t, y_i^t,z_i^t)$ is obtained via $\begin{bmatrix} x_i^t & y_i^t & z_i^t & 1 \end{bmatrix}^T  =\delta_{u_i^t,v_i^t} \cdot \mathbf{P}^{-1}\begin{bmatrix} u_i^t & v_i^t & 1 \end{bmatrix}^T,$
%
%
where $\mathbf{P}$ is the camera intrinsic matrix, and $\delta_{u_i^t,v_i^t}$ is the relative depth at $(u_i^t,v_i^t)$ obtained by~\cite{Ranftl2022}.
The spatial constraint $f_s$ is formulated as 
 %
 \begin{equation}
    f_s(\mathbf{p}^t_i,\mathbf{p}^t_j) = \mathbb{I}[d(\mathbf{p}^t_i,\mathbf{p}^t_j)\leq \mu],
    \label{eq:4}
\end{equation}
where $\mathbf{p}^t_i$ denotes the 3D coordinate of $i$-th object at time $t$, $\mathbb{I}(\cdot)$ is the indicator function, $d(\mathbf{p}^t_i,\mathbf{p}^t_j)$ computes the Euclidean distance between object $i$ and object $j$ in the 3D space, and $\mu$ is the relative distance threshold.
The motivation of spatial constraint is that interactions between two distant objects are usually scarce. 
{In our implementation, we empirically set the value of $\mu$ to be 3.0.} 

\subsubsection{Ego-Stuff Graph}\label{subsection:ego-stuff graph}
An \textit{Ego-Stuff Graph} $\mathbf{G}^{ES}$ is constructed in a similar manner as an \textit{Ego-Thing Graph} $\mathbf{G}^{ET}$ except for node feature extraction.

\noindent{\textbf{Node feature extraction.} }
We define the following classes as \textit{Stuff} objects: \textit{Crosswalk}, \textit{Lane Markings}, \textit{Lane Separator}, \textit{Road}, \textit{Service Lane}, \textit{Traffic Island}, \textit{Traffic Light} and \textit{Traffic Sign}. 
Some \textit{Stuff} classes (e.g., crosswalk obtained from semantic segmentation) cannot be well depicted in rectangular bounding boxes.
Thus, RoIAlign \cite{HeCVPR2017} is not applicable.
We propose MaskAlign to extract features from a binary mask $\mathbf{M}_i^{t}$, i.e., the $i$-th \textit{Stuff} object at time $t$. 
MaskAlign first downsamples the mask $\mathbf{M}_i^{t}$ to the same spatial dimension of an intermediate I3D \cite{CarreiraCVPR2017} feature map $\mathbf{X}$.
Note that the downsampled mask is denoted as ${\mathbf{M}_i^{t}}'$.
A \textit{Stuff} object feature is obtained as follows:
 \begin{equation}
    \mathbf{x}_i^{t} = \frac{\sum_{w=1}^{W}\sum_{h=1}^{H} \mathbf{X}^{t}_{(w,h)} \cdot {{\mathbf{M}_{i}^{t}}'}_{(w,h)} }{\sum_{w=1}^{W}\sum_{h=1}^{H} {{\mathbf{M}_{i}^{t}}'}_{(w,h)}},
\end{equation}
where $\mathbf{X}^{t}_{(w,h)} \in \mathbb{R}^{1\times D}$ is a D-dimension feature at location $(w,h)$ for time $t$, and ${{\mathbf{M}_{i}^{t}}'}_{(w,h)}$ is a binary scalar indicating whether object $i$ exists at location $(w,h)$.

\noindent{\textbf{Relational Modeling.}} We neglect interactions among \textit{Stuff} objects since by definition, \textit{Stuff} objects are presumed to not have interactions with each other.
We only model interactions between \textit{Stuff} objects and \textit{Ego}.
Hence, we set $f_s$ (as in Eq.~\ref{eq:4}) to 0 for every pair of \textit{Stuff} objects.
To model spatial constraint, we unproject every pixel within a downsampled binary mask ${{\mathbf{M}_{i}^{t}}'}$ to the 3D space, and calculate the relative distance between the corresponding 3D coordinates and the 3D coordinate of \textit{Ego}.
We choose the minimum distance within a downsampled mask. 
The distance threshold $\mu$ in \textit{Ego-Stuff Graphs} is empirically set to be 0.6.

\subsubsection{Interaction Modeling as Message Passing}
In Sec.\ref{subsection:ego-thing graph} and~\ref{subsection:ego-stuff graph}, two relational modelings are discussed.
To predict driver response, we need interaction modeling that captures influences of multiple traffic participants on a driver.
We formulate interactions as message passing in GCN that takes a graph as input, passes information through edges, and outputs updated nodes' features.
The message passing process in GCN is written as:

\begin{equation}
 \mathbf{X}^{l+1} = \mathbf{G}\mathbf{X}^{l}\mathbf{W}^{l}+\mathbf{X}^{l},
\end{equation}
where $\mathbf{G}$ is the affinity matrix discussed in Sec.\ref{subsection:ego-thing graph} and~\ref{subsection:ego-stuff graph}.
The matrix $\mathbf{X}^{l} \in \mathbb{R}^{(K+1)\times D}$ is the appearance feature matrix for the $l$-th layer.
$\mathbf{W}^{l} \in \mathbb{R}^{ D \times D}$ is a trainable weight matrix. 
We also build a residual connection by adding $\mathbf{X}^{l}$. 
Layer Normalization \cite{BaArxiv2016} and ReLU are applied before $\mathbf{X}^{l+1}$ is fed to the next message passing.
Note that we use a one-layer GCN to model \textit{Ego-Stuff} interactions and a two-layer GCN for \textit{Ego-Thing} interaction modeling.

\begin{figure}
    \centering
    \includegraphics[width=\columnwidth]{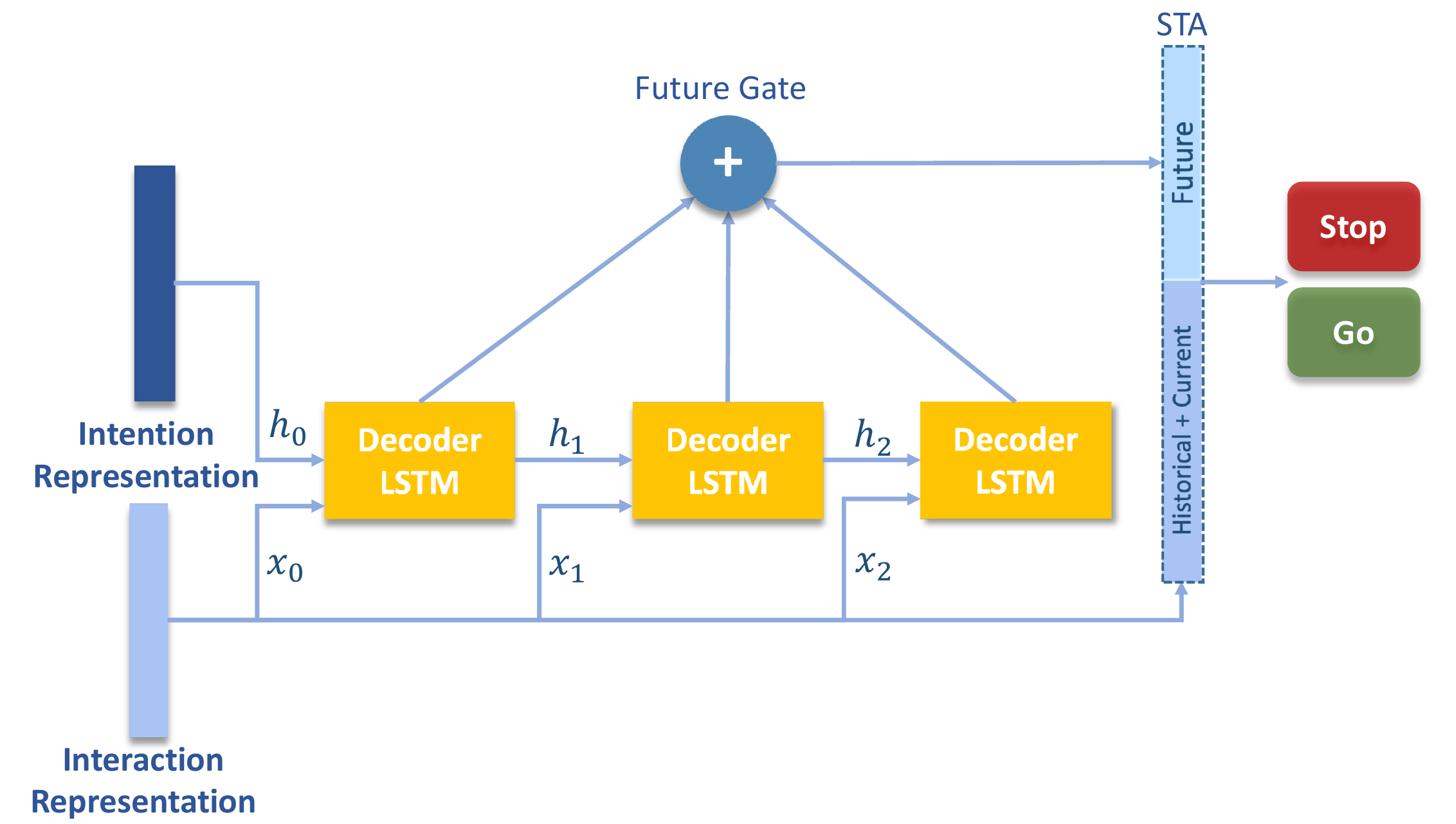}
    \caption{Decoder Temporal Recurrent Network (TRN) \cite{Xu_TRN_iccv2019}. The inputs to this module are intention and interaction representations. Note that intention representation is used to initialize the hidden state of the first decoder LSTM cell. The future gate and spatiotemporal accumulator (STA) aggregate features from historical, current, and predicted future information to predict driver response.}
    \label{fig:trn}
\end{figure}

\subsection{Projection}
\label{subsec:projection}
The role of \textit{Projection} is to forecast future actions of elements in the environment.
The updated appearance feature $\mathbf{X}^{l+1}$, discussed in the \textit{Comprehension} section, is processed independently at every frame without considering temporal changes. 
An encoder-decoder architecture is proposed to capture temporal interactions for forecasting future interactions. 

\noindent{\textbf{Encoder-decoder Architecture.}}
We implement the proposed  encoder-decoder architecture based on the Temporal Recurrent Network (TRN)~\cite{Xu_TRN_iccv2019}, which makes use of both accumulated historical evidence and predicted future information to better predict current action.
Following \cite{Xu_TRN_iccv2019}, we use LSTM~\cite{hochreiter1997long} as the backbone for both encoder and decoder.

We aggregate updated \textit{Ego} features from \textit{Ego-Stuff Graphs} and \textit{Ego-Thing Graphs} by an element-wise summation.
Time-specific updated \textit{Ego} features are fed into the encoder LSTM to obtain a $1\times D$ feature vector called interaction representation.
Note that prior works \cite{Wang_videospacetime_eccv2018, Wu_groupGCN_cvpr2019,Yan_stgcn_aaai2018} fuse all nodes' features in a graph, and the fused features are sent to the encoder LSTM. 
In contrast, we only send updated \textit{Ego} features in $\mathbf{X}^{l+1}$ to the encoder-decoder architecture, because updated \textit{Ego} features are expected to capture interactions among traffic participants that are key to robust driver response prediction. 
Unlike typical decoder architectures implemented as other LSTMs, TRN includes an LSTM decoder, a future gate, and a spatiotemporal accumulator (STA).
We extend TRN for the predicting driver response, and the corresponding architecture is depicted in Fig.~\ref{fig:trn}.
The LSTM decoder learns a feature representation of the evolving interactions.
The future gate receives a vector of hidden states (more details in Sec.\ref{section:implementation_details}) from the decoder LSTM and embeds features via the element-wise summation as the future context. 
The STA concatenates historical, current, and predicted future spatiotemporal features, and estimates driver response occurring in the very next frame.

\noindent{\textbf{Intention-aware Design.}}
Driver intention is indispensable for planning the next action \cite{codevilla2018end}, estimating the importance of road users\cite{rahimpour2019context}, and assessing risk \cite{lefevre2013intention}.
Similarly, in our task, driver response (i.e., \textit{Go} and \textit{Stop}) is determined not only by interactions among traffic participants but also driver intention (e.g., \textit{Left Turn} or \textit{Right Turn}).
For instance, a vehicle turning right at an intersection will not stop for pedestrians walking on the left sidewalk.
Hence, we treat features extracted from the I3D head as the intention representation, {since I3D features capture the historical motion dynamics and imply intention information}.
The representation is used to initialize the hidden state of the first decoder LSTM cell.
Note that the design differs from \cite{Xu_TRN_iccv2019}, which initializes the hidden state, $h_0$, with zeros.

\section{Causal Reasoning}
The previous section introduces the proposed driver behavior model. 
In this section, we discuss how we utilize \textit{intervention}, a powerful tool for causal inference, as a means for data augmentation to improve the performance of the driver behavior model (Sec.~\ref{subsection:training}) and apply causal inference to identify the risk object (Sec.~\ref{subsection:causal_inference}).

\subsection{Driver Behavior Model Training with Data Augmentation via Intervention}
\label{subsection:training}

\begin{figure*}[h!]
    \centering
    \includegraphics[width=\textwidth]{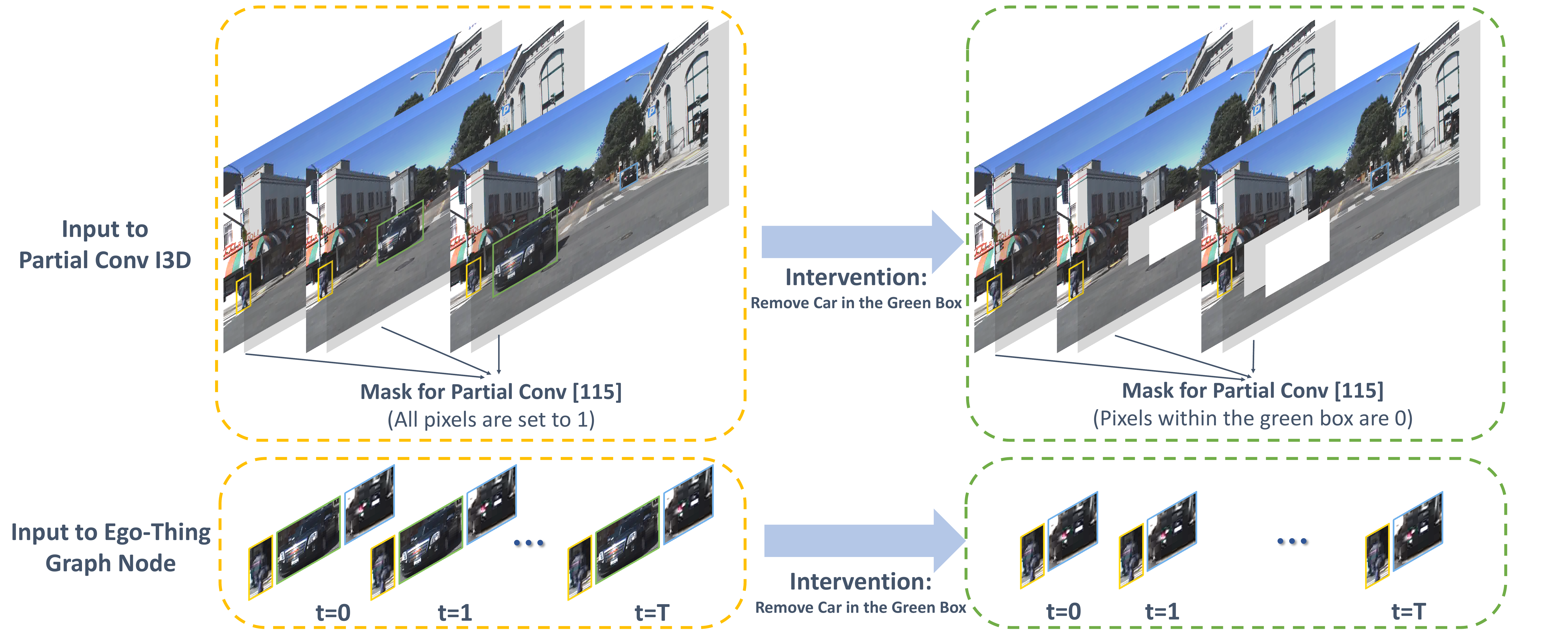}
    \caption{
    We simulate a situation using partial convolutional layers~\cite{liu2018image}.
    Note that a partial convolutional layer is initially introduced for image inpainting. 
    We utilize partial convolutions to simulate a scenario without the  presence of an object.
 The left-hand side of the figure depicts when an intervention is disabled.
 To simulate a situation without an object (e.g., the car in the green box), we set the pixels of the binary mask within the green box to 0. 
 In addition, the \textit{Ego-Thing Graph} is constructed without considering this car as a node.
 }
    \label{fig:input_data}
\end{figure*}

\begin{algorithm}[t]

	\caption{Driver Behavior Model Training}
	\begin{algorithmic}
	    \State $T$: Number of frames 
    	\State $N$: Number of \textit{Thing} objects in a given tracklet list

	    \State $\textbf{A}_r$: Ground truth driver response (either \textit{Go} or \textit{Stop})
	    \State \textbf{Input}: A sequence of RGB frames $I \coloneqq \{I_1, I_2,\cdots, I_T\}$

        \State \textbf{Output}: Predicted driver response $\textbf{a}_r$ and intention $\textbf{a}_i$. Notice that $\textbf{a}_r$ consists of confidence scores of \textit{Go} or \textit{Stop}. $\textbf{a}_r \coloneqq \{r^{go}, r^{stop}\}$. 
	    \algrule
	\end{algorithmic}
	
		\begin{algorithmic}[1]
	    \State $O 
         \begin{aligned}[t]
            &  \coloneqq\texttt{DetectionAndTracking($I$) } \\
            & \coloneqq \{O_1, O_2, \cdots, O_N\}   \text{ // List of \textit{Thing} object tracklets } 
         \end{aligned}$  
        \State $S 
         \begin{aligned}[t]
            &  \coloneqq\texttt{SemanticSegmentation($I$) } \\
            & \coloneqq \{S_1, S_2, \cdots, S_T\}   \text{ // List of \textit{Stuff} objects } 
         \end{aligned}$  
        
        \State $\begin{aligned}[t] &\text{// \textbf{Data Augmentation via Intervention (Sec.~\ref{subsection:training})}}
         \end{aligned}$
        \If{$\textbf{A}_r$ is \textit{Go} and $N > 1$}
            \State $\begin{aligned}[t] &\text{// Randomly remove a tracklet} \\
            & k \coloneqq \texttt{RandomSelect}(N)
            \end{aligned}$
        \Else
            \State $k$ is empty
        \EndIf
            
        \State $\begin{aligned}[t] &\text{// Mask out \textit{Thing} object $k$ on each mask frame} \\ 
            & M \coloneqq \texttt{MaskGenerator}(I, O_k)
            \end{aligned}$ 
        
        \State $ \begin{aligned}[t] &\text{// Remove a \textit{Thing} object $k$ from the tracklet list} \\
        & O' = O - \{O_k\}
        \end{aligned}$ 
        
        \State $\textbf{a}_r, \textbf{a}_i \coloneqq \texttt{DrivingModelTraining}(I,M,O',S)$ 
        //Discussed in Sec. \ref{subsection:training}
        \State \Return $ \textbf{a}_r, \textbf{a}_i $
	\end{algorithmic} 

	\label{algo1}
\end{algorithm}

{The performance of learning-based driver behavior models depends in a large part on the amount of training data under different traffic configurations~\cite{XuCVPR2017}.}
We propose a novel data augmentation strategy via \textit{intervention}~\cite{Pearl_causality_2009}. 
{Intervention is a means to differentiate among the different causal structures that are compatible with an observation~\cite{hagmayer2007causal}. 
The different causal structures between two events A and B are either A causes B, or B causes A, or they do not influence each other but they have a common cause.
%
%
We assume that non-causal objects do not influence the behavior of the driver. Thus, we can generate a new data point using the concept of intervention, i.e., removing non-causal objects.}
For instance, in a \textit{Go} scenario, a driver enters an intersection while pedestrians walk on the sidewalk in an opposite direction.
It is reasonable to assume that driver behavior is the same if a pedestrian is not present.
By removing the pedestrian, we can generate a new data point for training driver behavior models.
Note that the proposed augmentation strategy is only applicable to \textit{Go} scenarios.
In \textit{Stop} scenarios, we need to know causal objects' locations to remove non-causal objects.
However, exhaustive risk object labeling is costly, and that is not the focus of this work.

We cannot remove causal objects and assume the corresponding driver response to be \textit{Go} even if causal objects are identified. This is because traffic situations are inherently complicated. The corresponding driver response is unclear when the causal objects are removed.
For instance, a driver is in a congestion situation (i.e., driver stops for the frontal vehicle), and the traffic light of the driver's lane is red.
In this situation, the frontal vehicle is labeled as the risk object (cause).
However, driver response remains the same if the frontal vehicle were not present because of the red light.
Generating \textit{Stop} scenarios is non-trivial, and we leave it for future works.

To train driver behavior models with the proposed data augmentation strategy, a model should be able to "intervene," i.e., remove a non-causal object from images. 
We realize the strategy by replacing standard convolutional layers in I3D with \textit{partial convolutional layers}~\cite{liu2018image, liu2018partial}.
Note that a partial convolutional layer is initially introduced for image inpainting.
We utilize partial convolutions to simulate a scenario without the presence of an object.
A 3D partial convolutional layer takes two inputs, i.e., a sequence of RGB frames and a one-channel binary mask for each frame.
The pixel values of a mask are 1 by default.
While training the driver behavior model with data augmentation, we set the pixels within the selected object to be 0.
In addition, the node of the selected object in a graph is disconnected from the rest of the objects.
Details can be found in Fig. \ref{fig:input_data}.

The proposed training process is outlined in Algorithm~\ref{algo1}.
Given training samples in a \textit{Go} scenario, we randomly select an object $k$ to intervene, i.e., simulating a situation without the presence of the object.
Specifically, given a tracklet $o_k$,
a one-channel binary mask $M_t$ at time $t$ is defined as 
\begin{equation}
 M_t(i, j) =
\begin{cases}
    0,& \text{if } (i, j) \text{ in region } o_k^t\\
    1,& \text{otherwise}
\end{cases} \\,
\end{equation}
where $o_k^t$ is the bounding box of the $k$-th object at time $t$, and $(i,j)$ is a pixel coordinate within the box.
Note that $k$-th object is discarded from the tracklet list while training the driver behavior model.

\begin{algorithm}[t]
	\caption{Causal Inference for DROID}
	\begin{algorithmic}
	    \State $T$: Number of frames 
    	\State $N$: Number of objects 
	    \State \textbf{Input}: A sequence of RGB frames $I \coloneqq \{I_1, I_2,\cdots, I_T\}$ where the ego vehicle stops
        \State \textbf{Output}: Risk object ID 
	    \algrule
	\end{algorithmic}
	\begin{algorithmic}[1]
	    \State $O 
         \begin{aligned}[t]
            &  \coloneqq\texttt{DetectionAndTracking($I$) } \\
            & \coloneqq \{O_1, O_2, \cdots, O_N\} \text{// List of \textit{Thing} object tracklets }
         \end{aligned}$  
         
         \State $S 
         \begin{aligned}[t]
            &  \coloneqq\texttt{SemanticSegmentation($I$) } \\
            & \coloneqq \{S_1, S_2, \cdots, S_T\}   \text{ // List of \textit{Stuff} objects } 
         \end{aligned}$  

        \For{$ O_k \in O $}
            \State $\begin{aligned}[t] &\text{// Mask out \textit{Thing} object $k$ on each frame} \\ 
            & M \coloneqq \texttt{MaskGenerator}(I, O_k)
            \end{aligned}$ 
             \State $\begin{aligned}[t] &\text{// Remove the \textit{Thing} object $k$ from the tracklet list}\\
             & O' = O - \{O_k\}
             \end{aligned}$
            \State $\begin{aligned}[t] & \text{// Predict driver response and intention}\\
            & \text{without the object $k$, where } \textbf{a}_r \coloneqq \{r^{go}_k, r^{stop}_k \}\\
            & \textbf{a}_r, \textbf{a}_i \coloneqq \texttt{DrivingModel}(I,M, O', S) \\
             \end{aligned}$
        \EndFor
    
    \State \Return $\arg\max \limits_{k}(r^{go}_k) $ 
	\end{algorithmic} 
	\label{algo3}
\end{algorithm}

\begin{table*}[t]
    \small
    \centering
        \begin{tabular}
            {@{}l @{\;}@{\;}c @{\;}@{\;} @{\;}@{\;}c @{\;}@{\;} @{\;} c @{\;} @{\;} c @{\;} @{\;} c @{\;}}
            \toprule
             \multirow{3}{*}{ \begin{tabular}{@{}c@{}} \\Model \end{tabular}} & 
             \multicolumn{4}{c}{Response}  & 
             \multicolumn{1}{c}{Intention}  \\
             \cmidrule(lr){2-5} \cmidrule(lr){6-6}
             &
             \begin{tabular}{@{}c@{}@{}}  Perplexity \\  \end{tabular} & 
             \begin{tabular}{@{}c@{}@{}}  Macro \\ Accuracy \end{tabular} & 
             \begin{tabular}{@{}c@{}@{}}  Micro \\ Accuracy \end{tabular} & 
             \begin{tabular}{@{}c@{}@{}}  Overall \\ mAP \end{tabular} & 
             \begin{tabular}{@{}c@{}@{}}  Overall \\ mAP   \end{tabular} \\
             \midrule
             1. I3D + LSTM             & 1.00  & 64.37 & 77.95 & 71.07 &  /    \\
             2. I3D + LSTM + Multi-head& 0.93  & 68.27 & 79.04 & 70.12 & 36.41 \\
             3. Pixel-level Attention\cite{kim2017interpretable}  & 0.89  & 76.15 & 80.21 & 78.57 &  /    \\
             4. Object-level Attention \cite{wang2018deep}& 0.84  & 78.81 & 83.19 & 79.02 &  /    \\   
            \midrule
             5. GCN                              & 0.83              & 77.57             & 82.64             & 80.33             &  /        \\
             6. GCN + Multi-head                & 0.72              & 76.30             & 85.68             & \underline{84.46}             &  36.31    \\
             7. GCN + TRN Head                   & \underline{0.69}  & \underline{79.32} & \underline{86.17} & 83.44 &  \textbf{36.80}     \\
             8. GCN + TRN Head + Data Augmentation    & \textbf{0.37}     & \textbf{87.63}    & \textbf{92.56}    & \textbf{95.44}    &  \underline{36.75}  \\             
            \bottomrule
        \end{tabular}
\vspace{8pt}
\caption{Results of our driver response prediction (rows 5-8) compared with baseline predictions (rows 1-4). Perplexity (lower is better), macro- and micro-average accuracies, and overall mAP are used as metrics for driver response prediction.
The unit is \% for all metrics except perplexity.
The best and second best performances are shown in bold and underlined, respectively.
We also report the performance of driver intention prediction using the overall mean average precision (mAP) as the metric.}
        \label{table:driving_model}
\end{table*}

\subsection{Causal Inference for DROID}\label{subsection:causal_inference}
Given a \textit{Stop} scenario, we aim to identify the corresponding risk object. 
We deploy the same intervention process discussed in Sec.\ref{subsection:training} to identify the risk object. 
Specifically, the masks of a tracklet and the corresponding video frames are processed by the same driver behavior model.
The model outputs the confidence score of \textit{Go} and \textit{Stop} without the presence of the tracklet.
After iterating through all tracklets, we select the object whose tracklet elimination yields the highest \textit{Go} confidence score to be the risk object.
This is because the object causes the most driver behavior change.
Algorithm~\ref{algo3} describes the overall causal inference process.

\section{Experiments}\label{section:experiments}

\subsection{Implementation Details}\label{section:implementation_details}
We implement our framework in TensorFlow. All experiments are performed on a server with four NVIDIA TITAN-XP cards.
The input to the framework is a 20-frame clip with a resolution of $224 \times 224$ at 3 fps, approximately 6.67 s.
The framework outputs the predictions of driver intention and response of the very next frame.
{For the data preprocess, we use \cite{rotabulo2017place} trained on Mapillary Vistas research set~\cite{neuhold2017mapillary} to perform semantic segmentation. We apply object detection to every frame via a Mask R-CNN model \cite{HeCVPR2017} trained on Cityscape dataset\cite{CordtsCVPR2016Cityscapes}.  Based on the detection results, Deep SORT \cite{Wojke2017simple} is utilized to associate detected objects into tracks. New objects are identified when the detection cannot be associated with an existing track.}

We adopt Inception-v3 \cite{Szegedy_inceptionv3_cvpr2016} pre-trained on ImageNet~\cite{Russakovsky_imagenet_ijcv2015} as the backbone, following \cite{CarreiraCVPR2017} to inflate 2D convolution into 3D convolution, and fine-tune it (i.e., I3D) on the Kinetics action recognition dataset \cite{Kay_kinetics_arvix2017}. 
The intermediate feature used in RoIAlign and MaskAlign is the \texttt{Mixed\_3c} layer, where the number of feature channels is 512. 
The intention feature is generated from a $1 \times 1 \times 1$ convolution on the \texttt{Mixed\_5c} layer's feature, and the channel number of the feature is 512.
%
The downsampled binary mask ${{\mathbf{M}_{i}^{t}}'}$ is $28\times 28$.
{The decoder length is set to 3, and the TRN model is identical to the original implementation \cite{Xu_TRN_iccv2019} except for the hidden state. The number of hidden cells we use is 1024 while it is 2\,000 in \cite{Xu_TRN_iccv2019}. Also, instead of initializing hidden states with zeros, we input the intention representation to fulfill our intention-aware design.}

The model is trained in a two-stage training scheme with a batch size of 32. 
{First, we finetune the Kinetics pre-trained I3D model on HDDS for 50\,000 iterations. At this stage, GCN and TRN are not involved, and we use the driver intention as the training target. }
{At the second stage, we employ the augmentation strategy mentioned in Sec.\ref{subsection:training}. The weights trained from the first stage are loaded to the completed driver behavior model with GCN and TRN. We finetune the network for another 20\,000 iterations. }
{We use the Adam optimizer~\cite{Kingma_adam_arvix2014} with the default parameters ($\beta_1$=0.9, $\beta_2$=0.999 and $\epsilon=1\times 10^{-8}$) for both stages.}
The learning rate is set to be 0.001 and 0.0002 for the first and second stages, respectively.

\subsection{Driver Behavior Model Performance}\label{section:driving_model_performance}
\subsubsection{Evaluation Setup}
The performance of the driver behavior model is evaluated as a discrete feasible action prediction, in accordance with~\cite{XuCVPR2017,wang2018deep,wang2019monocular}.
The two discrete actions, \textit{Go} and \textit{Stop}, are evaluated. 
We follow the train/test1 split in HDDS, where 846\,411 and 271\,989 samples are used for training and testing, respectively.
Four evaluation metrics are utilized. 
First, we report perplexity as in~\cite{XuCVPR2017,wang2018deep,wang2019monocular}.
Perplexity calculates the negative log-likelihood of predicted probability of \textbf{Response} given ground truth (lower is better).
Second, we report the macro-averaged accuracy.
Note that, in a multi-class classification setup, the micro-averaged accuracy is preferable if the label distribution is imbalanced.
In our task, the \textit{Go} to \textit{Stop} ratio is approximately 4:1. 
Therefore, we also report the micro-averaged accuracy as the third metric.
\textbf{Response} prediction can be treated as an online action detection task~\cite{de2016online,Xu_TRN_iccv2019}.
We use per-frame mean average precision (mAP) as the fourth evaluation metric.

\subsubsection{Evaluation} \label{sec:compare_with_state_of_art}

Table~\ref{table:driving_model} summarizes the results of the driver behavior models.
We compare the following baselines.
To compare different models, we keep their backbone network (i.e., Inception-v3) the same.

\noindent\textbf{I3D+LSTM.}  We extract visual features from the \texttt{Mixed\_5c} layer of I3D and sequentially input the features at each time step to a two-layer LSTM~\cite{Xu_TRN_iccv2019} for temporal modeling.

\noindent\textbf{Pixel-level attention.} The pixel-level attention module is proposed by~\cite{kim2017interpretable} to improve model's intepretability and performance.

\noindent\textbf{Object-level attention.} In \cite{wang2018deep}, the authors propose an object-centric attention mechanism to augment end-to-end policy learning. 
Both pixel- and object-level attention modules are incorporated into \textbf{I3D+LSTM}.

The following summarizes our proposals.

\noindent\textbf{GCN.} {This method is based on our previous work \cite{Chengxi2020}.} The key difference between GCN and three baselines is the input feature to the LSTM module.
Specifically, the feature is processed via GCNs and contains interaction between traffic participants and driver.

\noindent\textbf{Multi-head.} We add an additional head for driver intention prediction to \textbf{I3D+LSTM} and \textbf{GCN}.
A standard cross-entropy loss is used for driver intention prediction.
Note that both the interaction and intention features share the same features from the \texttt{Mixed\_5c} layer of I3D.

\noindent\textbf{TRN Head.} To forecast future interactions, we incorporate TRN~\cite{Xu_TRN_iccv2019}.
%
We initialize TRN with intention representation (as shown in Fig.~\ref{fig:trn}) .

\noindent\textbf{Intervention.} The concept of intervention is utilized to augment training data to improve the performance of driver response prediction discussed in Sec.\ref{subsection:training}.
We show that \textbf{GCN} outperforms baselines, demonstrating the importance of interaction modeling.
By incorporating \textbf{Multi-head}, i.e., intention modeling, both extensions reduce the perplexity by 0.07 and 0.11, respectively.
With \textbf{TRN Head}, we observe that perplexity is reduced by 0.03.
Finally, we demonstrate that \textbf{Intervention} significantly improves the performance of the driver response prediction (0.32 decrease in perplexity). 
More detailed discussion can be found in Sec.\ref{section:ablation_study_1}.

While promising improvements are observed for driver response prediction, the trend does not hold for driver intention prediction, as shown in Table~\ref{table:driving_model}.
{Our hypothesis is that this is because driver response prediction utilizes TRN with the interaction and intention representations.
Note that the interaction representation is obtained via extracting features from the \texttt{Mixed\_3c} layer of I3D, processing the features via GCN, and encoding those processed features by LSTM.
In contrast, driver intention prediction only uses the intention representation obtained from the I3D head after the \texttt{Mixed\_5c} layer. 
Therefore, the intention representation is not trained effectively due to its architectural design.
%
In the future work, we plan to incorporate road structures~\cite{Wang_top_CVPR2019,Philion_lift_eccv2020,Tian_roadgraph_icra_2021} or explicitly model possible goals~\cite{Mangalam_iccv2021} to improve the performance of driver intention prediction.}

\begin{table}[t]
    \small
    \centering
        \begin{tabular}
            {@{}l l@{} c@{}}
            \toprule
              & Model &  Perplexity \\
            \midrule
            \multirow{3}*{ \begin{tabular}{@{}l@{}} Intention \\ Modeling \end{tabular}} & Without intention modeling & 0.83 \\
            & Multi-head & 0.72 \\
            & TRN Head & \textbf{0.69} \\
           \midrule
           \multirow{3}*{ \begin{tabular}{@{}l@{}}Different \\ Graphs\end{tabular}}  
           & Ego-Stuff Graph & 0.74 \\
            &Ego-Thing Graph & 0.80 \\
            &Ego-Thing Graph + Ego-Stuff Graph & \textbf{0.69} \\
            \midrule
             \multirow{2}*{ \begin{tabular}{@{}l@{}}Spatial \\ Modeling \end{tabular}}&Appearance Relation  &  0.73 \\
            &Appearance + Spatial Relation & \textbf{0.69} \\
            \midrule
             \multirow{2}*{ \begin{tabular}{@{}l@{}} Data \\ Augmentation \end{tabular}}& Without Augmentation & 0.69 \\
            & With Augmentation  &  \textbf{0.37} \\
            \bottomrule
        \end{tabular}
        \vspace{8pt}
        \caption{Ablative study of our design choices. A lower perplexity indicates better model performance. The best-performing models within each category (row labels) are shown in bold.
        }

        \label{table:ablation_studies_driving_model}
\end{table}
\subsubsection{Ablation Study}\label{section:ablation_study_1}
We conduct ablation studies to understand the contributions of the proposed architecture designs. The studies are summarized in Table \ref{table:ablation_studies_driving_model}.

\noindent\textbf{Analysis of Intention Modeling.}
{The first section of Table~\ref{table:ablation_studies_driving_model} analyzes the importance to perplexity of intention modeling. 
The baseline does not consider intention.
When intention representation is incorporated into \textbf{Multi-head} and \textbf{TRN Head}, the perplexity is improved by 0.11 and 0.14, respectively.
The experiment empirically shows that driver intention is indispensable for planning the next action, as mentioned in Sec.\ref{subsec:projection}.
{\textbf{TRN Head} performs slightly better than \textbf{Multi-head}. This could be because \textbf{TRN Head} leverages both the historical and predicted future interactions with other traffic participants for driver response prediction, while \textbf{Multi-head} design does not have the advantage of future prediction. We empirically demonstrate the effectiveness of the \textbf{TRN Head} design.}
}

\noindent\textbf{Variations of Different Graphs.}
{The experiments aim to prove the importance of modeling interactions with both \textit{Thing} and \textit{Stuff} for driver response prediction. From Table~\ref{table:ablation_studies_driving_model}, we observe that \textit{Ego-Stuff} and \textit{Ego-Thing Graphs} capture different aspects of interactions with ego-vehicles. When both \textit{Ego-Stuff} and \textit{Ego-Thing} interactions are considered jointly, the model achieves the best perplexity results.
The results empirically show the hypothesis that an explicit modeling of drivers, traffic participants, and road infrastructure is crucial for driver response prediction
}

\noindent\textbf{Importance of Spatial Relation.}
{We study the importance of the spatial relation function (Eq. \ref{eq:4}) to the \textbf{Response} prediction.
We conduct two experiments, i.e., 1) using only the appearance relations, and 2) appending 3D spatial relation as an additional constraint. 
As we live in a 3D world, an interaction model for driver response prediction should take 3D spatial relations into account. 
As shown in Table~\ref{table:ablation_studies_driving_model}, we confirm our hypothesis that the response prediction performance is superior when the proposed 3D spatial constraint is utilized.
}

\noindent\textbf{Data Augmentation via Intervention.}
{We study the impact of data augmentation by comparing the performance of two models trained with and without the data augmentation strategy.
%
%
In particular, augmentation plays a vital role in learning-based solutions for various computer vision tasks.
In this work, we utilize the concept of \textit{intervention} to synthesize new training data. Note that we also use \textit{intervention} for risk object identification. 
The last section in Table~\ref{table:ablation_studies_driving_model} showcases the significance of using augmented data, which cuts the perplexity by nearly half.
The data augmentation strategy adds variations to the training set that improve the robustness of the driver behavior model.
}

\begin{table}[t]
    \small
    \centering
    \resizebox{\columnwidth}{!}{
        
        \begin{tabular}
            {@{}c c@{\;}@{\;}c@{\;}@{\;}c@{\;}@{\;}c@{\;}@{\;}c@{}}
            
        \toprule
         \multirow{3}{*}{ \begin{tabular}{@{}c@{}} \\Model \end{tabular}} & \multicolumn{4}{c}{$mAcc$} \\
        \cmidrule(lr){2-5}
        & \begin{tabular}{@{}c@{}}Crossing\\Vehicle \end{tabular}
        & \begin{tabular}{@{}c@{}}Crossing\\Pedestrian\end{tabular}
        & \begin{tabular}{@{}c@{}}Parked\\Vehicle \end{tabular}
        &\begin{tabular}{@{}c@{}} Congestion \end{tabular}\\
        \midrule
        
       \footnotesize  Random Selection  & \ 15.1  & 7.1 & 6.4 & 5.5 \\
       \footnotesize  Driver's Attention Prediction *~\cite{Xia_ACCV_2018}    & 16.8 & 8.9 & 10.0  &21.3 \\
       \midrule
       \footnotesize Object-level Attention *~\cite{wang2018deep} & 22.6 & 9.5 & 22.6 & 40.7 \\
       \footnotesize Pixel-level Attention *~\cite{kim2017interpretable} & 28.0 & 8.1 & 15.6  & 35.7 \\
        \midrule
         \footnotesize GCN  (ours)    & 27.5 & \textbf{13.6} & 26.0 & 51.3 \\
        \footnotesize  GCN + TRN Head (ours)  & \underline{29.0} & \underline{13.2} & \underline{27.3} & \underline{52.2} \\
        \footnotesize  GCN + TRN Head + Data Augmentation (ours) & \textbf{32.5} & 12.9 & \textbf{28.4} & \textbf{57.5} \\   
    \bottomrule
    \end{tabular}
    }
    \vspace{8pt}
    \caption{Comparison with baselines. 
    %
    %
    The methods with * are re-implemented by us to ensure the
    same backbone is used for fair comparisons. 
    $mAcc$ stands for mean accuracy, and the unit is \%. 
    The best and second best performances are shown in bold and underlined, respectively.
    }
    \label{table:comparison_baselines}
\end{table}

\begin{table*}[t]
    \small
    \centering
    \vspace{5pt}
    \resizebox{\textwidth}{!}{
        
        \begin{tabular}
            {@{} c   c  @{\;} c  @{\;} 
            @{\;} c @{\;}   @{\;} c @{\;}    @{\;} c  @{\;}
            @{\;} c @{\;}   @{\;} c @{\;}    @{\;} c  @{\;}
            @{\;} c @{\;}   @{\;} c @{\;}    @{\;} c  @{\;} 
            @{\;} c @{\;}   @{\;} c @{\;}    @{\;} c  @{}}
            
        \toprule
         \multirow{3}{*}{ \begin{tabular}{@{}c@{}}Driver Behavior Model\end{tabular}}  &  \multirow{3}{*}{ \begin{tabular}{@{}c@{}} Data \\ Augmentation \end{tabular} } &  \multirow{3}{*}{ \begin{tabular}{@{}c@{}} Causal \\ Inference \end{tabular} }&\multicolumn{3}{c}{Crossing Vehicle} &   \multicolumn{3}{c}{Crossing Pedestrian} & \multicolumn{3}{c}{Parked Vehicle} & \multicolumn{3}{c}{Congestion}  \\
        \cmidrule(lr){4-6} \cmidrule(lr){7-9}\cmidrule(lr){10-12}\cmidrule(lr){13-15}
             &
             &
             &
            \begin{tabular}{@{}c@{}} $Acc_{0.5}$ \end{tabular} &
            \begin{tabular}{@{}c@{}} $Acc_{0.75}$\end{tabular} &
            \begin{tabular}{@{}c@{}}$mAcc$\end{tabular} &
            \begin{tabular}{@{}c@{}}$Acc_{0.5}$\end{tabular} &
            \begin{tabular}{@{}c@{}} $Acc_{0.75}$\end{tabular} &
            \begin{tabular}{@{}c@{}} $mAcc$\end{tabular} &
            \begin{tabular}{@{}c@{}}$Acc_{0.5}$\end{tabular} &
            \begin{tabular}{@{}c@{}}$Acc_{0.75}$\end{tabular} &
            \begin{tabular}{@{}c@{}}$mAcc$\end{tabular} &
            \begin{tabular}{@{}c@{}} $Acc_{0.5}$\end{tabular} &
            \begin{tabular}{@{}c@{}}$Acc_{0.75}$\end{tabular} &
            \begin{tabular}{@{}c@{}}$mAcc$\end{tabular}\\

            \cmidrule{1-15}
            I3D + LSTM  & \xmark  &\cmark&  29.9    &   29.9    &   26.3    &   15.5 &	14.3 &	12.4& 33.1 &	28.7 &	25.4& 39.4 &	35.4 &	32.9 \\
            \cmidrule{1-15}
             GCN (ours) & \xmark &\cmark& 31.8  &	31.5 &	27.5 & \underline{16.7} &	15.5 &	\underline{13.6} & 32.4 &	29.4 &	26.0 & 56.6 &	56.6	& 51.3 \\
             GCN + Multi-head (ours) & \xmark &\cmark& 31.8 &	31.8 &	28.0 & \textbf{17.9} & \textbf{17.9} &	\textbf{14.6}  & 32.4 &	29.4 &	26.3 & 61.6 & 57.6 & 53.8 \\
             GCN + TRN Head (ours) & \xmark &\cmark& \underline{33.1}	& \underline{33.1} &\underline{29.0} & \underline{16.7} &	\underline{16.7} &	13.2 & \underline{33.8} &	\underline{30.2} &	\underline{27.3} &  60.6 &	56.6 &	52.2  \\
             GCN + TRN Head (ours) & \cmark &\xmark& 28.3 &	28.0&	25.0 & 13.1 &	11.9 &	9.6&	22.1&	21.3  & 18.7 & \underline{65.7}&	\underline{61.6} & \underline{57.4}\\
             
              GCN + TRN Head (ours)& \cmark &\cmark& \textbf{37.0} &	\textbf{37.0} &	\textbf{32.5} & 15.5 &	15.5 &	12.9 & \textbf{35.3} &	\textbf{31.6} &	\textbf{28.4}  & \textbf{66.7} &	\textbf{62.6} &	\textbf{57.5} \\

\bottomrule
        \end{tabular}
    }
    \vspace{8pt}
    \caption{Ablation study of the proposed framework. The unit is \%. The best and second best performances are shown in bold and underlined, respectively. Mean accuracy ($mACC$) and accuracies at IoU thresholds of 0.5 ($mACC_{0.5}$) and 0.75 ($mACC_{0.75}$) are reported.}
    \label{table:cause_object_identification_acc}
\end{table*}

\subsection{Driver-centric Risk Object Identification}
\subsubsection{Evaluation Setup}\label{sec:ROI_eval_setup}
We evaluate DROID in the four reactive scenarios: \textit{Congestion}; \textit{Crossing  Pedestrian}; \textit{Crossing Vehicle}; and \textit{Parked Vehicle}.
We use accuracy (number of correct predictions over the number of samples) as the evaluation metric.
A correct prediction is one that has an Intersection over Union (IoU) score between a selected box and a ground truth box that is larger than a predefined threshold.
Similar to~\cite{lin2014microsoft,zhang2019self}, accuracies at IoU thresholds of 0.5 and 0.75 are reported.
In addition, mean accuracy ($mACC$) is calculated by using IoU thresholds ranging from 0.5 to 0.95 (in increments of 0.05).

\subsubsection{Evaluation}
We compare the performance of DROID with the following baselines.
The results are shown in Table~\ref{table:comparison_baselines}.

\noindent\textbf{Random Selection.}
Random selection randomly picks an object as the risk object from all the detections for a given frame.
Note that the method does not process any visual information except by means of object detection.
The method is used to contextualize the challenge of this task.

\noindent\textbf{Driver Attention Prediction} uses a pre-trained model~\cite{Xia_ACCV_2018} trained on the (Berkeley DeepDrive) BDD-A dataset to predict the driver's gaze attention maps at each frame.
We compute an average attention weight of every detected object region based on a predicted attention map.
The risk object is the object with the highest attention weight, indicating the driver's gaze attends to this region.
The model is trained with human gaze signals that are unavailable in HDD.
The performance of this method is slightly better than \textbf{Random Selection}, as reported in the second row of  Table~\ref{table:comparison_baselines}.
We observe that predicted attention maps tend to focus at a vanishing point.
Note that this issue has been raised in~\cite{Tawari_salient_2018}, highlighting the problem as one of the challenges of imitating human gaze behavior.

\noindent\textbf{Object-level Attention Selector}. 
The object-level attention driving model~\cite{wang2018deep} is reformulated for DROID. The risk object is the object with the highest object-attention score. 

\noindent\textbf{Pixel-level Attention}.
Kim et al.~\cite{kim2017interpretable} propose a causality test to search for regions that influence the network's output behavior.
Note that region proposals are formed based on sampling predicted pixel-level attention maps. 
To identify a risk object, 
we replace the region proposal strategy used in~\cite{kim2017interpretable} with object detection, and utilize the inferred pixel-level attention map to filter out detections with low attention values. 
In the experiments, we set the threshold at 0.002.
The modification ensures a fair comparison, as region proposals obtained from~\cite{kim2017interpretable} are not guaranteed to be an object entity. 
Note that the code of region proposal generation detailed in~\cite{kim2017interpretable} is not publicly available.

We report favorable DROID performance over existing baselines~\cite{Xia_ACCV_2018}, \cite{kim2017interpretable,wang2018deep} in Table~\ref{table:comparison_baselines}. 
The results indicate the effectiveness of the proposed driver behavior model and causal inference for the task.
In the next section, we perform ablation studies to examine the contributions of each part of our model.
Notice that our evaluation protocol differs from \cite{li2020make}. 
In \cite{li2020make}, the authors train four different models and test four scenarios independently, whereas a single driver behavior model is trained in this work.

\begin{figure*}[t!]
\subfloat[Crossing Vehicle \label{fig:risk_object_identification_a}]{
\minipage{0.246\textwidth}
\minipage{\linewidth}
  \includegraphics[page=1,width=\linewidth]{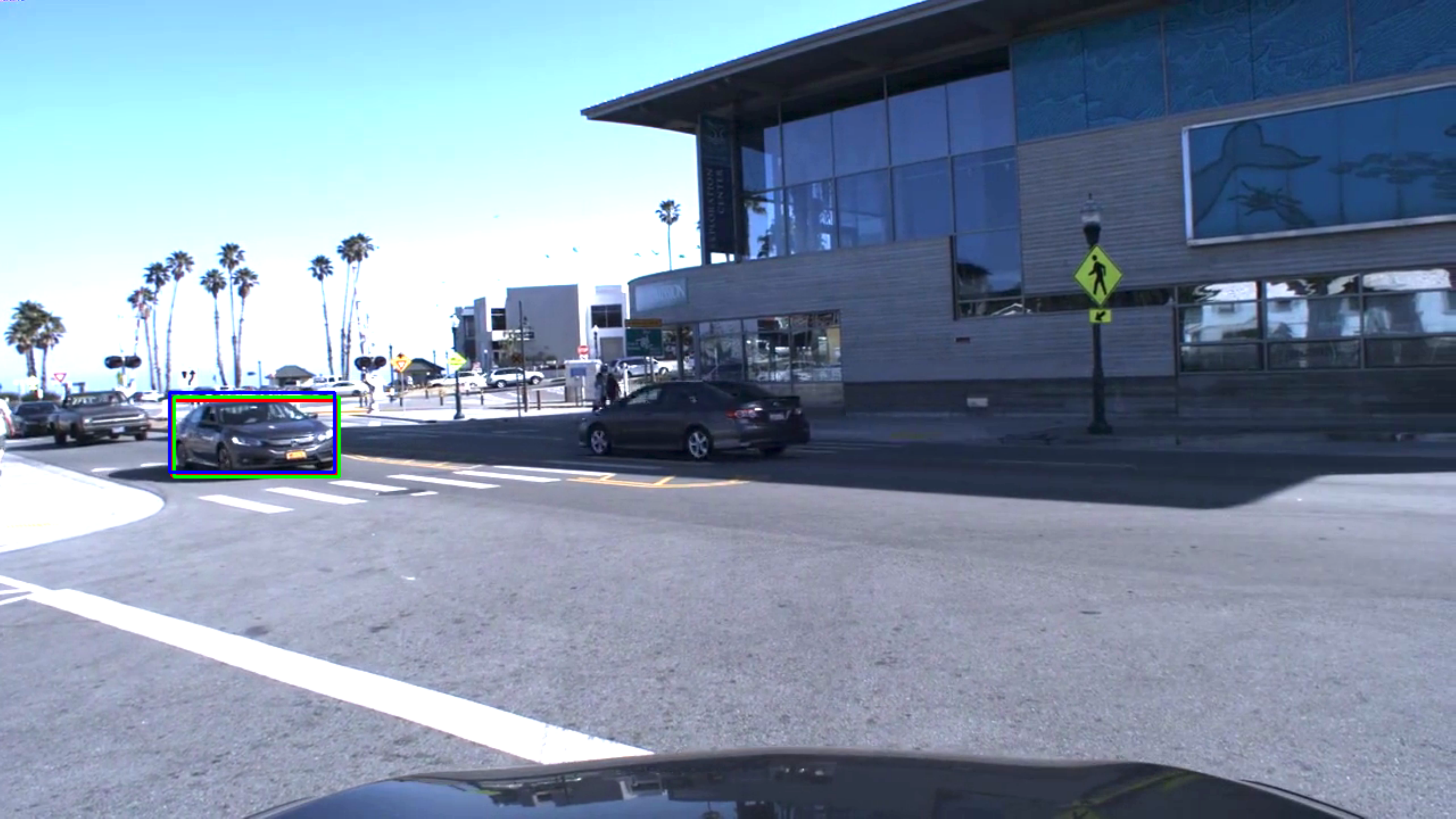}
\endminipage\hfill
\minipage{\linewidth}
  \includegraphics[page=1,width=\linewidth]{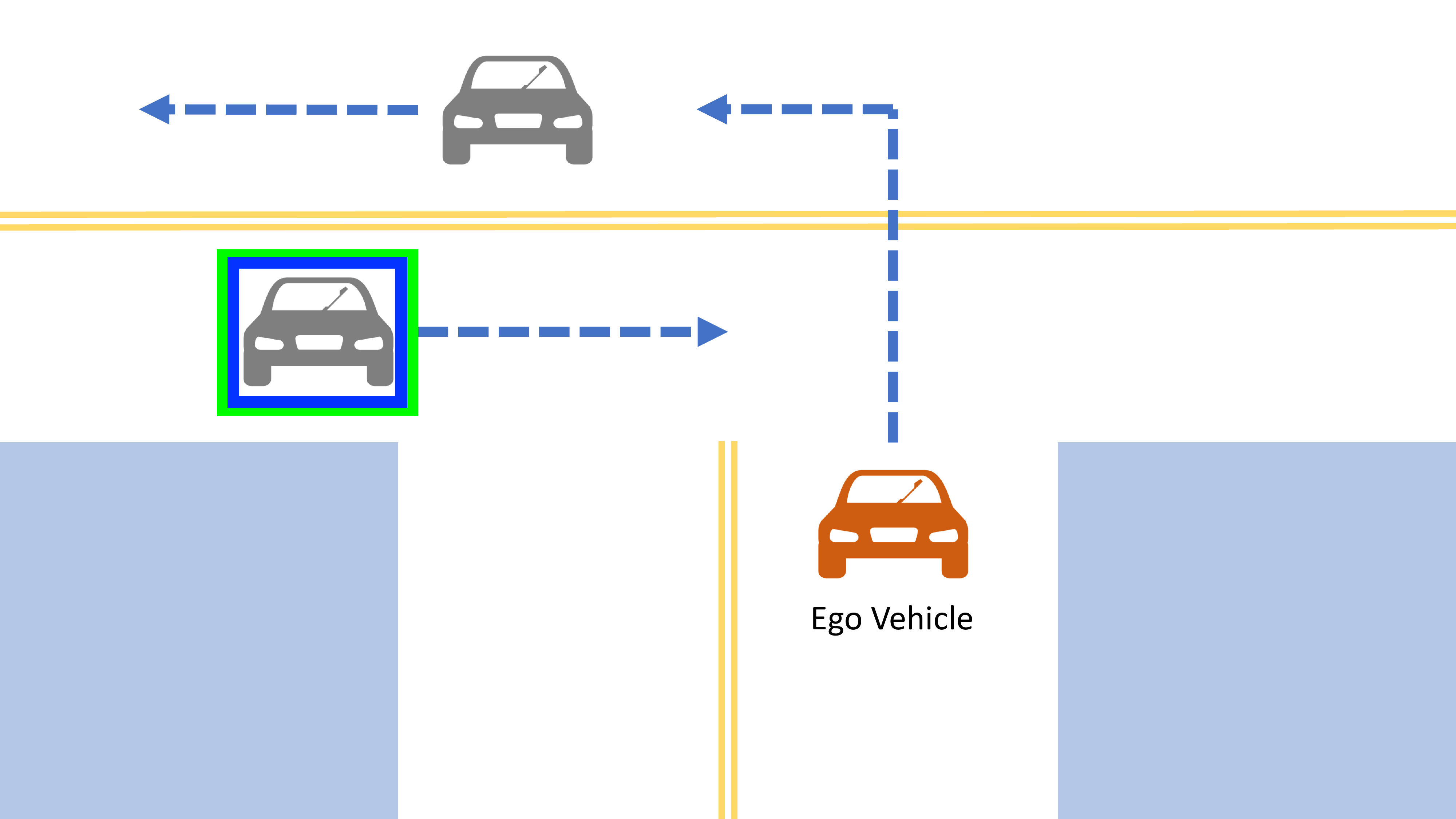}
\endminipage\hfill
\endminipage
}
\subfloat[Crossing Pedestrian \label{fig:risk_object_identification_b}]{
\minipage{0.246\textwidth}
\minipage{\linewidth}
  \includegraphics[page=2,width=\linewidth]{figures/success_example/success_example.pdf}
\endminipage\hfill
\minipage{\linewidth}
  \includegraphics[page=2,width=\linewidth]{figures/success_example/BEV.pdf}
\endminipage\hfill
\endminipage
}
\subfloat[Parked Vehicle \label{fig:risk_object_identification_c}]{
\minipage{0.246\textwidth}
\minipage{\linewidth}
  \includegraphics[page=3,width=\linewidth]{figures/success_example/success_example.pdf}
\endminipage\hfill
\minipage{\linewidth}
  \includegraphics[page=3,width=\linewidth]{figures/success_example/BEV.pdf}
\endminipage\hfill
\endminipage
}
\subfloat[Congestion \label{fig:risk_object_identification_d}]{
\minipage{0.246\textwidth}
\minipage{\linewidth}
  \includegraphics[page=4,width=\linewidth]{figures/success_example/success_example.pdf}
\endminipage\hfill
\minipage{\linewidth}
  \includegraphics[page=4,width=\linewidth]{figures/success_example/BEV.pdf}
\endminipage\hfill
\endminipage
}
\hfill
\subfloat[Crossing Vehicle \label{fig:risk_object_identification_e}]{
\minipage{0.246\textwidth}
\minipage{\linewidth}
  \includegraphics[page=5,width=\linewidth]{figures/success_example/success_example.pdf}
\endminipage\hfill
\minipage{\linewidth}
  \includegraphics[page=5,width=\linewidth]{figures/success_example/BEV.pdf}
\endminipage\hfill
\endminipage
}
\subfloat[Crossing Pedestrian \label{fig:risk_object_identification_f}]{
\minipage{0.246\textwidth}
\minipage{\linewidth}
  \includegraphics[page=6,width=\linewidth]{figures/success_example/success_example.pdf}
\endminipage\hfill
\minipage{\linewidth}
  \includegraphics[page=6,width=\linewidth]{figures/success_example/BEV.pdf}
\endminipage\hfill
\endminipage
}
\subfloat[Parked Vehicle \label{fig:risk_object_identification_g}]{
\minipage{0.246\textwidth}
\minipage{\linewidth}
  \includegraphics[page=7,width=\linewidth]{figures/success_example/success_example.pdf}
\endminipage\hfill
\minipage{\linewidth}
  \includegraphics[page=7,width=\linewidth]{figures/success_example/BEV.pdf}
\endminipage\hfill
\endminipage
}
\subfloat[Congestion \label{fig:risk_object_identification_h}]{
\minipage{0.246\textwidth}
\minipage{\linewidth}
  \includegraphics[page=8,width=\linewidth]{figures/success_example/success_example.pdf}
\endminipage\hfill
\minipage{\linewidth}
  \includegraphics[page=8,width=\linewidth]{figures/success_example/BEV.pdf}
\endminipage\hfill
\endminipage
}
\hfill
\subfloat{
\minipage{\textwidth}
\includegraphics[page=1,width=\linewidth]{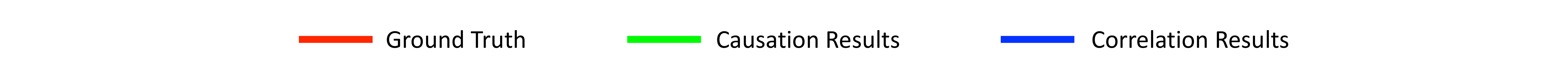}
\endminipage
}
  \caption{DROID results obtained by \textbf{Causation} and \textbf{Correlation}.
  Note that \textbf{Causation} is the causal inference-based approach proposed in the paper.
  Instead of using causal inference, \textbf{Correlation} determines the risk object by selecting the object sending the highest attention weight to \textit{Ego} in the \textit{Ego-Thing Graph}.
  The top row shows an egocentric view where green boxes indicate our \textbf{Causation} results, blue boxes are \textbf{Correlation} results, and ground truth boxes are in red. 
  A bird's-eye-view representation is presented in the bottom row, providing information including scene layout and intentions of traffic participants.}
\label{fig:risk_object_identification}
\end{figure*}

\begin{figure*}[t!]
\subfloat[Crossing Vehicle\label{fig:risk_score_a}]{
\minipage{0.54\columnwidth}
  \includegraphics[width=\linewidth,trim = {0 0 0 0},clip]{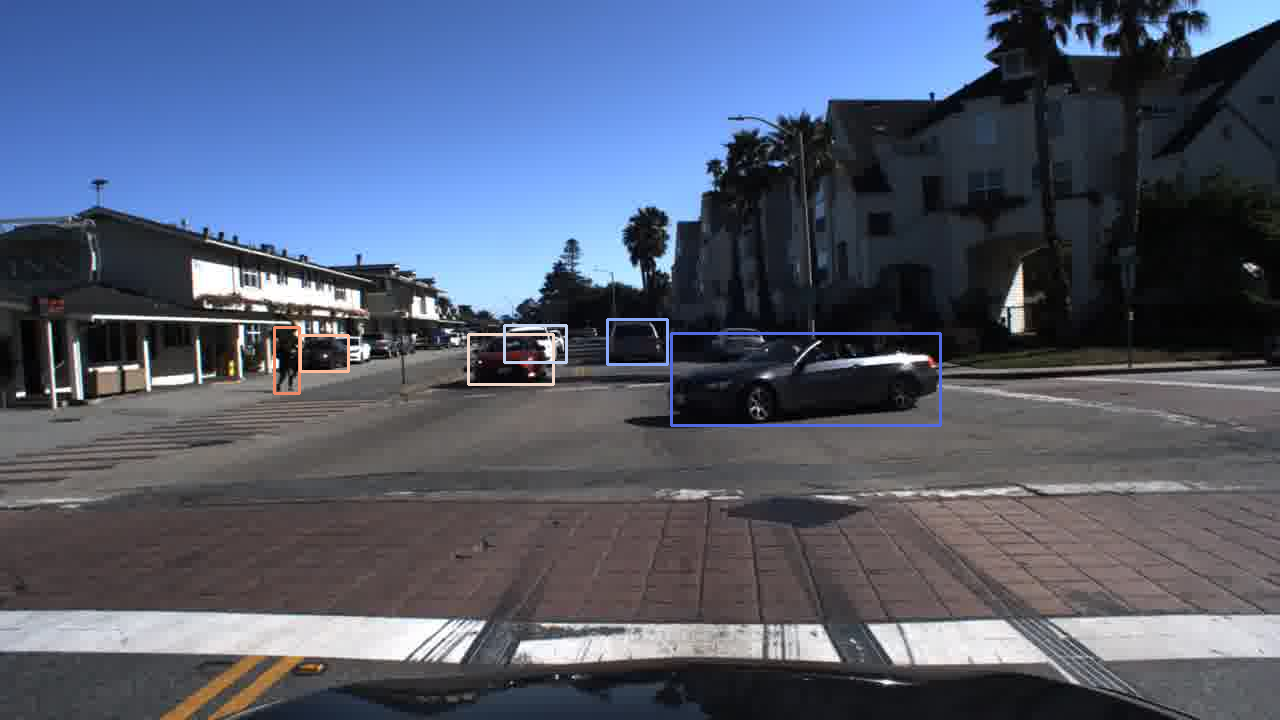}
\endminipage
\minipage{0.46\columnwidth}
  \includegraphics[width=\linewidth,trim = {18 18 18 18},clip]{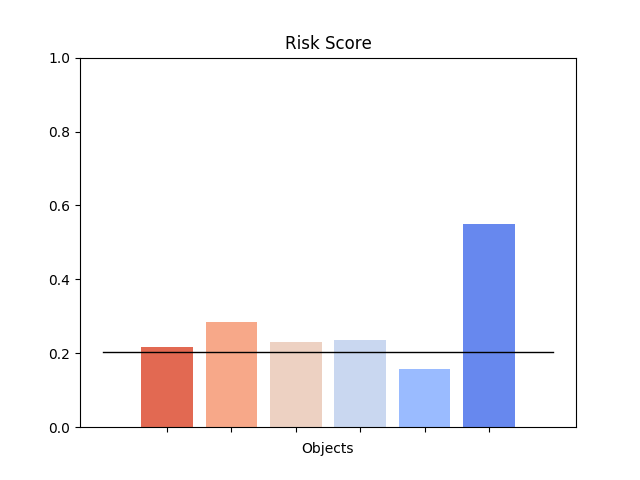}
\endminipage}
\subfloat[Crossing Pedestrian\label{fig:risk_score_b}]{
\minipage{0.54\columnwidth}
  \includegraphics[width=\linewidth,trim = {0 0 0 0},clip]{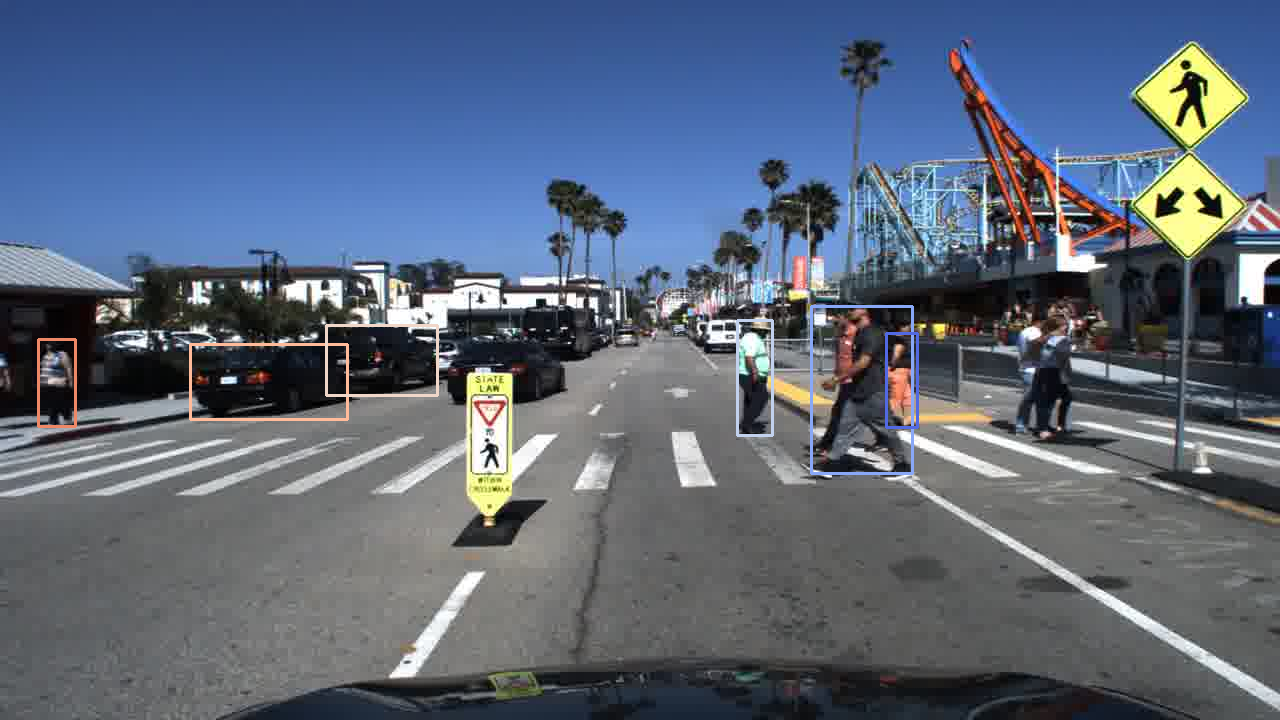}
\endminipage
\minipage{0.46\columnwidth}%
  \includegraphics[width=\linewidth,trim = {18 18 18 18},clip]{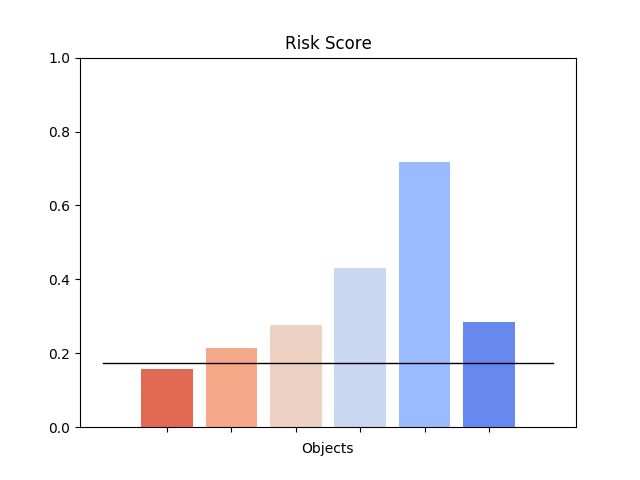}
\endminipage 
}
\hfill
\subfloat[Parked Vehicle\label{fig:risk_score_c}]{
\minipage{0.54\columnwidth}
  \includegraphics[width=\linewidth,trim = {0 0 0 0},clip]{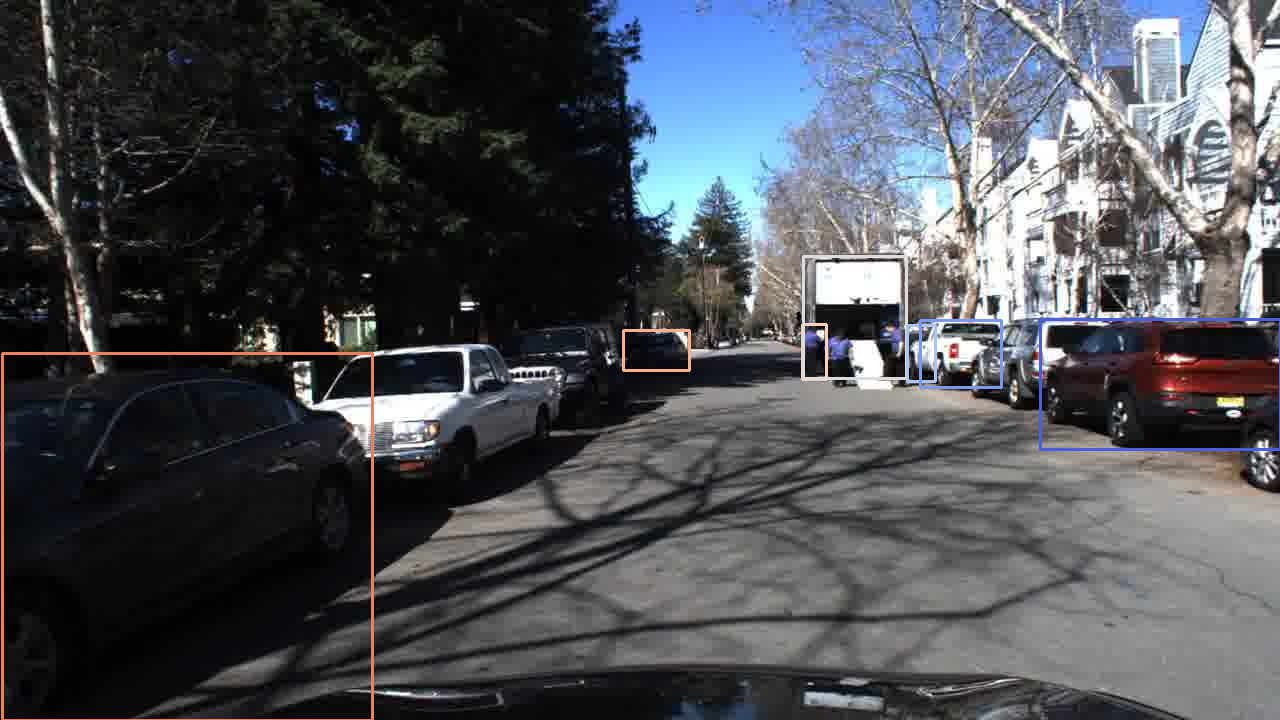}
\endminipage
\minipage{0.46\columnwidth}
  \includegraphics[width=\linewidth,trim = {18 18 18 18},clip]{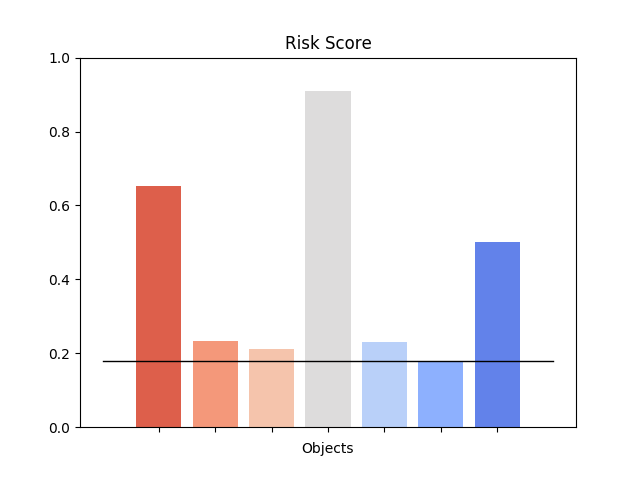}
\endminipage}
\subfloat[Congestion \label{fig:risk_score_d}]{
\minipage{0.54\columnwidth}
  \includegraphics[width=\linewidth,trim = {0 0 0 0},clip]{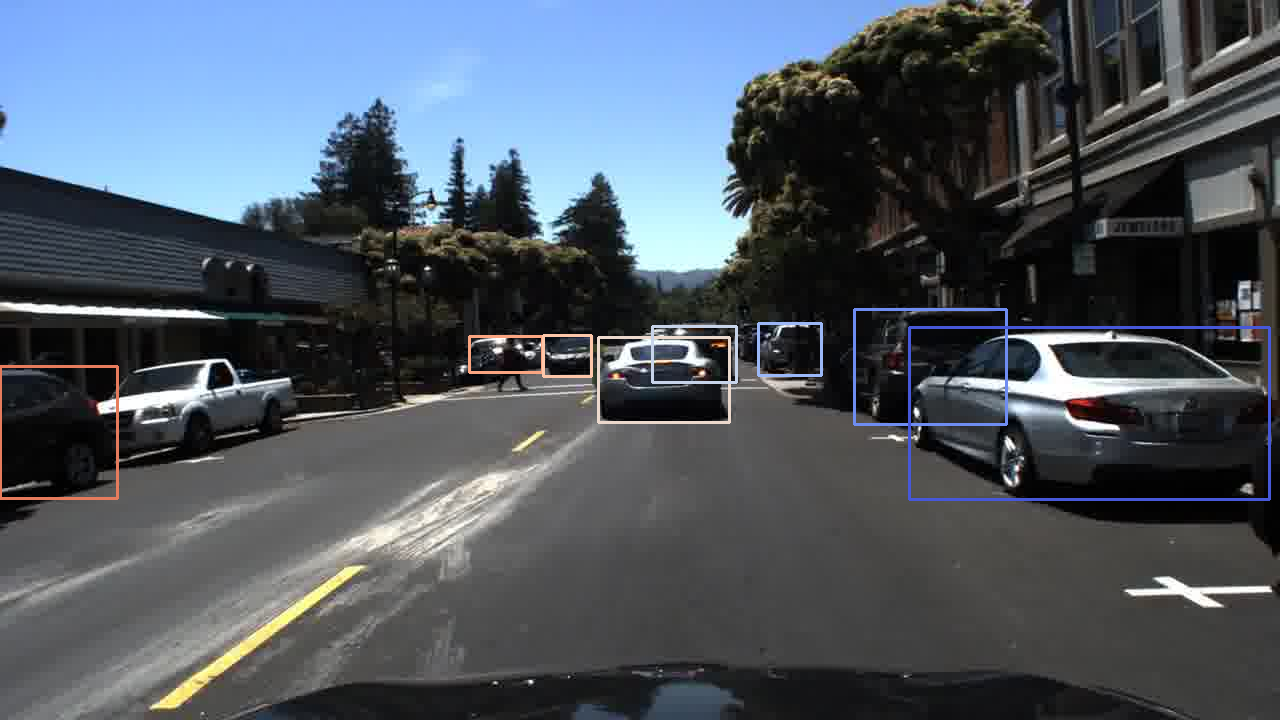}
\endminipage
\minipage{0.46\columnwidth}
  \includegraphics[width=\linewidth,trim = {18 18 18 18},clip]{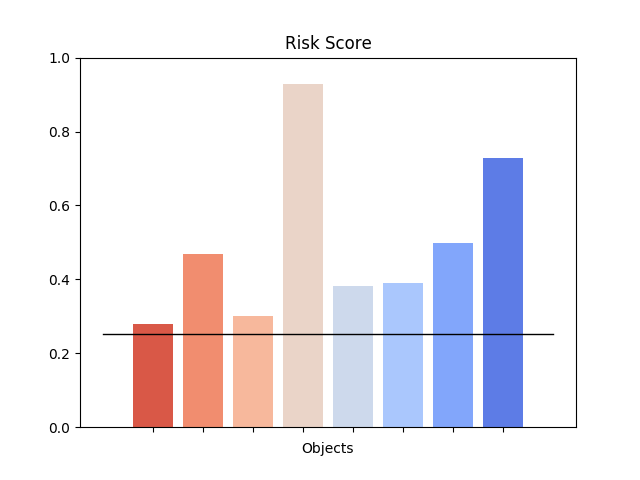}
\endminipage}

\caption{
%
Risk Assessment Visualization. 
On the left, all detected objects are shown in bounding boxes with different colors. The risk score of each object is depicted in a bar chart on the right. The color of each bar is one-to-one matched to the bounding box. We use a black horizontal line to indicate the predicted \textit{Go} score of the driver without applying any intervention.
}
\label{fig:risk_score}
\end{figure*}

\subsubsection{Ablation Study}
Three variations are studied to analyze their impacts on the performance of DROID: (1) architecture of the driver behavior model, (2) intention modeling and (3) training strategy.
The results are summarized in Table~\ref{table:cause_object_identification_acc}.

\noindent\textbf{Architecture}. The completed framework (GCN + TRN Head, reported in the last row of Table~\ref{table:cause_object_identification_acc}) boosts the $mACC$s of GCN by $6.2\%$, $0.5\%$, $3.0\%$ and $24.6\%$ in four different scenarios, respectively. 
This architecture ranks first in three senarios (\textit{Crossing Vehicle}, \textit{Parked Vehicle}, and \textit{Congestion}).
In most scenarios, using GCN architecture performs better than the pure I3D + LSTM model. These results show a similar trend in the performance of driver response prediction in Table~\ref{table:driving_model} (Row 5 v.s. Row 1, and Row 6 v.s. Row 2). The results are aligned with the design of the causal inference-based framework, i.e., the identification of risk object(s) relies on the prediction of driver response when different objects are removed. A better driver response prediction improves the performance of risk object identification. 

\noindent\textbf{Intention Modeling}. 
Multi-head and TRN head-based intention modelings improve the accuracy of identifying risk objects. This observation also aligns with common senses: the risk object varies depending on different navigation goals, i.e. intentions. In Section\ref{sec:compare_with_state_of_art}, the TRN head-based approach achieves a better performance for the driver response prediction task than Multi-head. However, we do not observe the same phenomenon in the DROID task. It could be because the two tasks are not designed in a unified manner. Thus, the value of TRN-based modeling cannot be observed with DROID. 

\noindent\textbf{Training with Data Augmentation}. We observe significant improvement in all scenarios with the proposed data augmentation strategy except for \textit{Crossing Pedestrian}.
The results indicate the effectiveness of the proposed training strategy.
For \textit{Crossing Pedestrian}, our conjecture is that vehicles are likely to be chosen as risk objects because of the natural imbalanced distribution in the training data.
Note that the ratio of detected vehicles to pedestrians is approximately 17:1.
Our model learns how to identify risk objects under traffic configurations (especially different vehicle configurations) so that the model performs favorably for scenarios that involve interacting with vehicles.
In contrast, scenarios that involve interacting with pedestrians are less emphasized.
To solve this problem, a possible solution is to perform a category-aware intervention so that a balanced distribution can be obtained.

In summary, with the proposed components, i.e., TRN Head, intention modeling, and training with data augmentation, we demonstrate state-of-the-art DROID performance.
We observe a similarly improved DROID performance in driver response prediction, discussed in Sec.\ref{section:driving_model_performance}.

\subsubsection{Correlation vs. Causation}
We study the importance of causal modeling for this task.
Instead of using causal inference (called \textbf{Causation}) to identify the risk object, the risk object is the object sending the highest attention weight to \textit{Ego} in \textit{Ego-Thing Graph}.
We call this method \textbf{Correlation}.
In Table~\ref{table:cause_object_identification_acc}, the second to the last row shows the results of \textbf{Correlation}.
Our \textbf{Causation} approach significantly outperforms \textbf{Correlation} in all reactive scenarios.
We empirically demonstrate the need of causal modeling for this task. 

In Fig.~\ref{fig:risk_object_identification}, ground truth risk objects are enclosed in red bounding boxes, our \textbf{Causation} results are shown in green, and the \textbf{Correlation} predictions are shown in blue boxes. 
In addition, we provide a bird’s-eye-view pictorial illustration of scenes in the second row.
Note that it depicts scene layouts, driver intention, and traffic participants' intentions, with identified risk objects in green boxes.
In Fig.~\ref{fig:risk_object_identification}b, three crossing pedestrians with different intentions are depicted.
Our \textbf{Causation} approach correctly identifies the left-hand side pedestrian as the risk object while the driver intends to turn left.
While \textbf{Correlation} predicts the same result, our method is more explainable because the decision is made by considering driver intention.  
Fig.~\ref{fig:risk_object_identification}d, f, g and h showcase examples where \textbf{Correlation} fails but \textbf{Causation} identifies risk objects successfully.

\subsection{Application: Risk Assessment}

{Our framework is able to perform risk assessment when multiple risk objects exist.}
%
We visualize objects' risk scores in Fig.~\ref{fig:risk_score} under different reactive scenarios.
All detected objects are encased in bounding boxes with different colors.
Their risk scores are shown in a bar chart with color.
The risk score of an object is equivalent to the predicted confidence score of \textit{Go} after the object is removed.
A higher confidence score of \textit{Go} means that the object has a higher chance of influencing driver behavior.
We use a black horizontal line to indicate the predicted confidence score of \textit{Go} when we do not intervene in the input.  
%
In all these cases, confidence scores are smaller than 0.5, representing correct driver response prediction by the proposed model.
Favorable risk assessment results are demonstrated. 
{In particular, in Fig.~\ref{fig:risk_score_b}, when the two pedestrians could potentially be risk objects, our framework assigns high risk scores to both and the pedestrian closer to the vehicle is rated with a higher risk score.}

{Due to iterative causal inference, our framework requires the same amount of computational time ($\sim$0.15 s) for each iteration with one NVIDIA TITAN-XP card. If there are more than 10 objects in a scene, the computation time of risk would be more than 1.5 s. Therefore, a single-shot design for risk assessment that does not require iterative causal inference is in need to realize future real-time applications of DROID.}


\section{Conclusion}
{In this paper, we focus on subjective risk assessment and operationalize the assessment by predicting driver behavior changes and identifying the cause of changes.} A new task called DROID, which uses egocentric video collected from front-facing cameras to identify object(s) influencing a driver’s behavior, given only the driver’s response as the supervision signal, is introduced. We formulate the task as a cause-effect problem. A novel two-stage framework inspired by the model of SA and causal inference is present. We also construct a dataset for DROID to evaluate the proposed system. Extensive quantitative and qualitative evaluations are conducted. Favorable performance compared with strong baselines is demonstrated. 
Future work can leverage road topology explicitly to improve driver intention prediction. 
Additionally, single shot risk assessment for DROID would be interesting to explore for
practical applications.

 
\section*{Acknowledgments}
The work is sponsored by Honda Research Institute USA. Yi-Ting Chen is supported in part by the Higher Education Sprout Project of the National Yang Ming Chiao Tung University and Ministry of Education (MOE), the Ministry of Science and Technology (MOST) under grants 110-2222-E-A49-001-MY3 and 110-2634-F-002-051, and Mobile Drive Technology Co., Ltd (MobileDrive).





\ifCLASSOPTIONcaptionsoff
  \newpage
\fi




\bibliographystyle{IEEEtran}
\bibliography{reference}

%




%
\vskip 0pt plus -1fil
\begin{IEEEbiography}[{\includegraphics[width=1in,height=1.25in,clip,keepaspectratio]{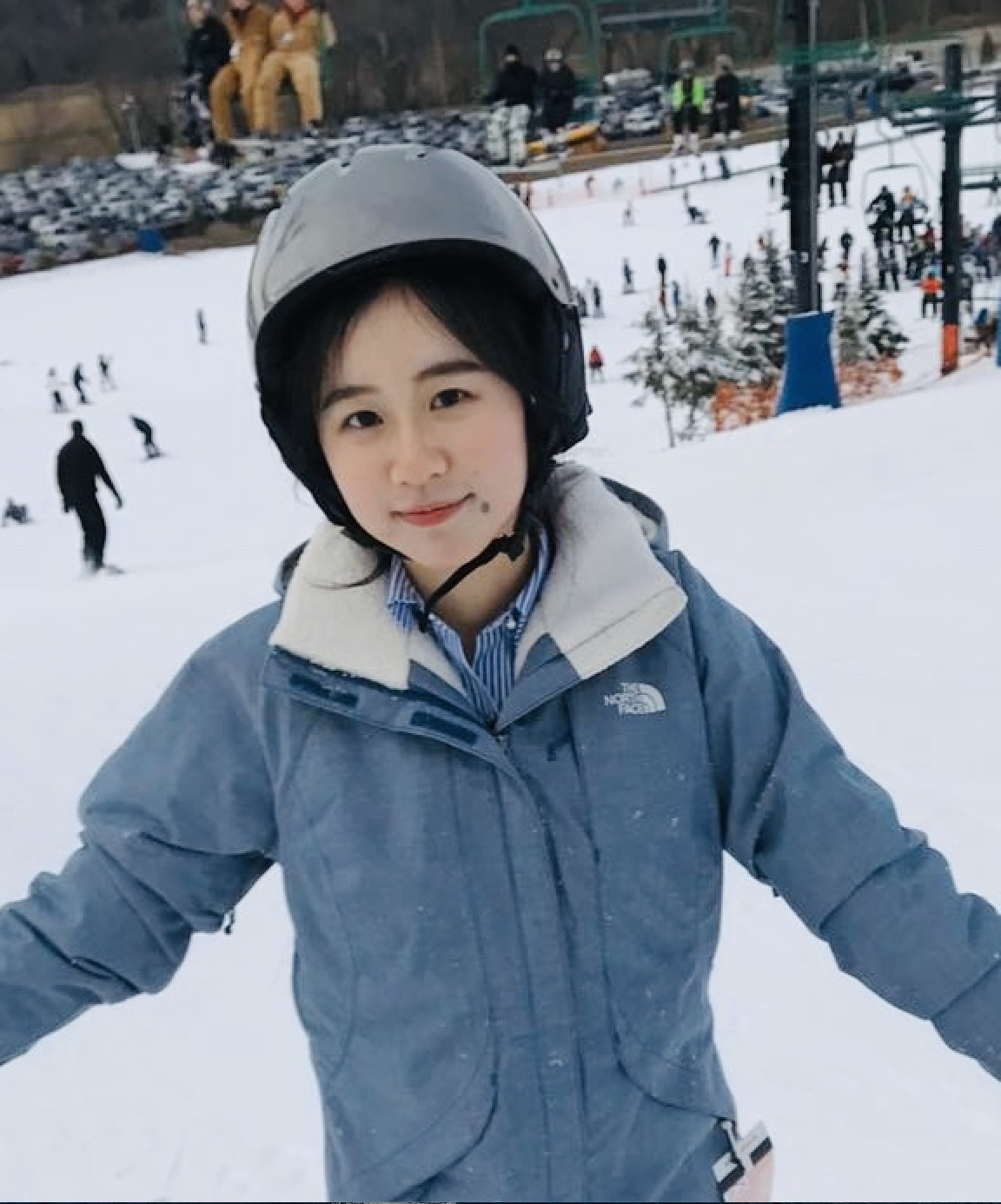}}]{Chengxi Li}
 received her B.S. degree (2016) in Electrical Engineering from Fudan University in Shanghai, China and a Ph.D. degree (2021) from the School of Electrical and Computer Engineering at Purdue University, IN, USA. She is currently a research scientist at Meta Platforms, Inc. Her research interests lie at the intersection of machine learning and computer vision. 
\end{IEEEbiography}
\vskip 0pt plus -1fil

\begin{IEEEbiography}[{\includegraphics[width=1in,height=1.25in,clip,keepaspectratio]{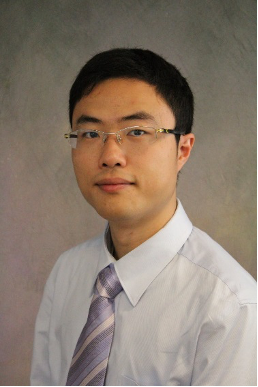}}]
{Stanley H. Chan}
(S'06--M'12--SM'17) received a B.Eng. degree (with first class honors) in Electrical Engineering from the University of Hong Kong in 2007, an M.A. degree in Mathematics from the University of California at San Diego in 2009, and a Ph.D. degree in Electrical Engineering from the University of California at San Diego in 2011. From 2012 to 2014, he was a postdoctoral research fellow at Harvard John A. Paulson School of Engineering and Applied Sciences. He is currently an associate professor in the School of Electrical and Computer Engineering and the Department of Statistics at Purdue University, West Lafayette, IN.

Dr. Chan is a recipient of the Best Paper Award of IEEE International Conference on Image Processing 2016, IEEE Signal Processing Cup 2016 Second Prize, Purdue College of Engineering Exceptional Early Career Teaching Award 2019, Purdue College of Engineering Outstanding Graduate Mentor Award 2016, and Eta Kappa Nu (Beta Chapter) Outstanding Teaching Award 2015. He is also a recipient of the Croucher Foundation Fellowship for Postdoctoral Research 2012-2013 and the Croucher Foundation Scholarship for Ph.D. Studies 2008-2010. His research interests include computational imaging and machine learning.

\end{IEEEbiography}
\vskip 0pt plus -1fil
\begin{IEEEbiography}[{\includegraphics[width=1in,height=1.25in,clip,keepaspectratio]{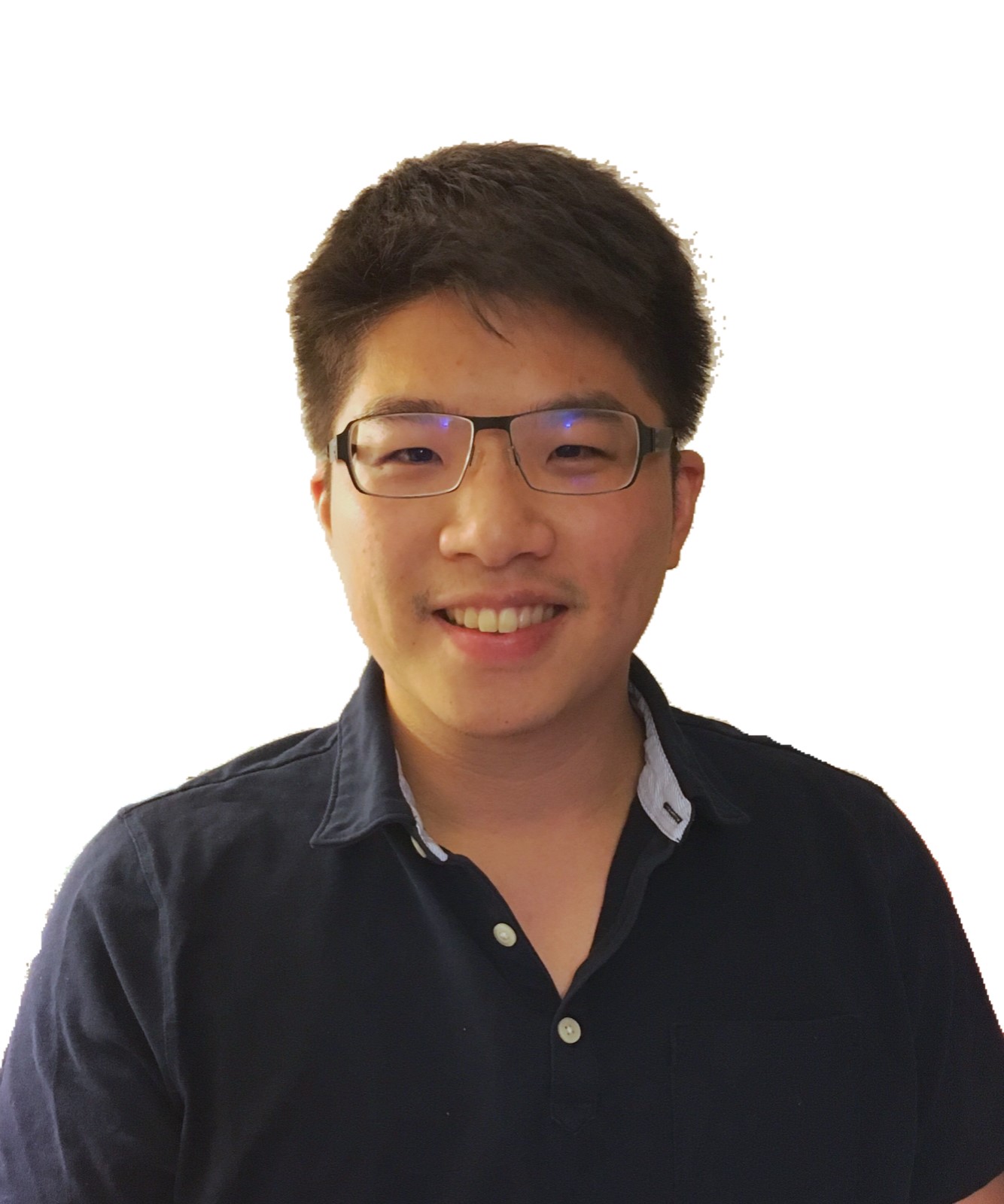}}]{Yi-Ting Chen}
received a B.S. degree in Electronics Engineering from National Chiao Tung University, Hsinchu, Taiwan and a Ph.D. degree from the Department of Electrical and Computer Engineering at Purdue University in 2015. He was a senior scientist at Honda Research Institute USA from 2015 to 2020. He is currently an assistant professor in the Department of Computer Science at National Yang Ming Chiao Tung University, Hsinchu, Taiwan. His research interests lie in computer vision, machine learning, robotics, and behavior science, and their applications to intelligent driving systems and assistive robotics.
\end{IEEEbiography}





\end{document}


\appendix

\begin{figure*}[ht]
\subfloat[ \textbf{Intention}: Intersection Passing \label{fig:visualization_scene_representation_a}]{
  \includegraphics[width=\columnwidth,page=1,trim={0 2cm 0 2cm },clip]{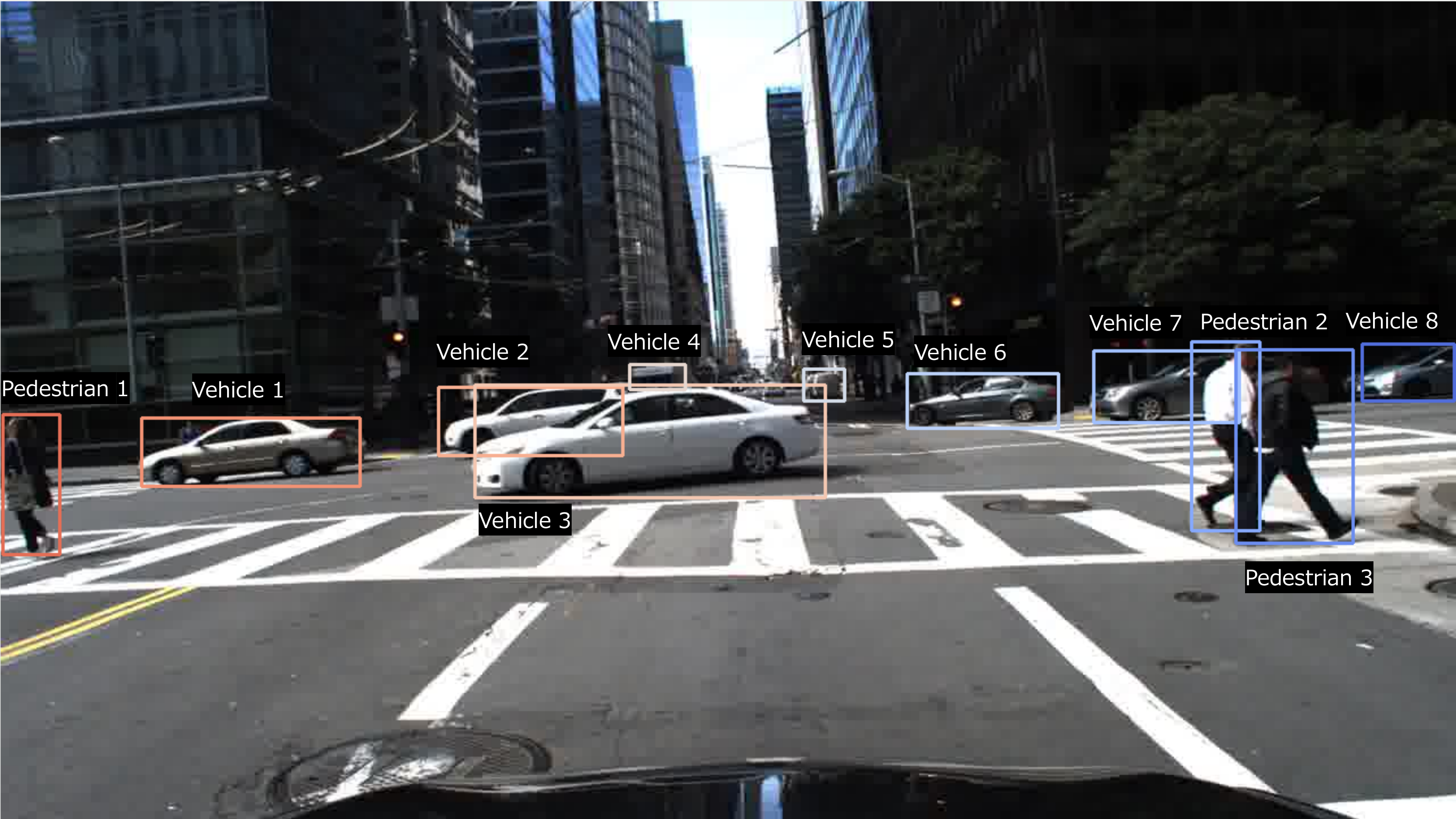}}
 \hfill
\subfloat[ \textbf{Intention}: Left Turn \label{fig:visualization_scene_representation_b}]{
  \includegraphics[width=\columnwidth,page=2,trim={0 2cm 0 2cm },clip]{figures/bev_graph/graph_vis_frame.pdf}}
 \hfill
\subfloat[Learned affinity matrix and BEV visualization.\label{fig:visualization_scene_representation_c}]{
\minipage{0.36\columnwidth}
  \includegraphics[width=\linewidth,page=1]{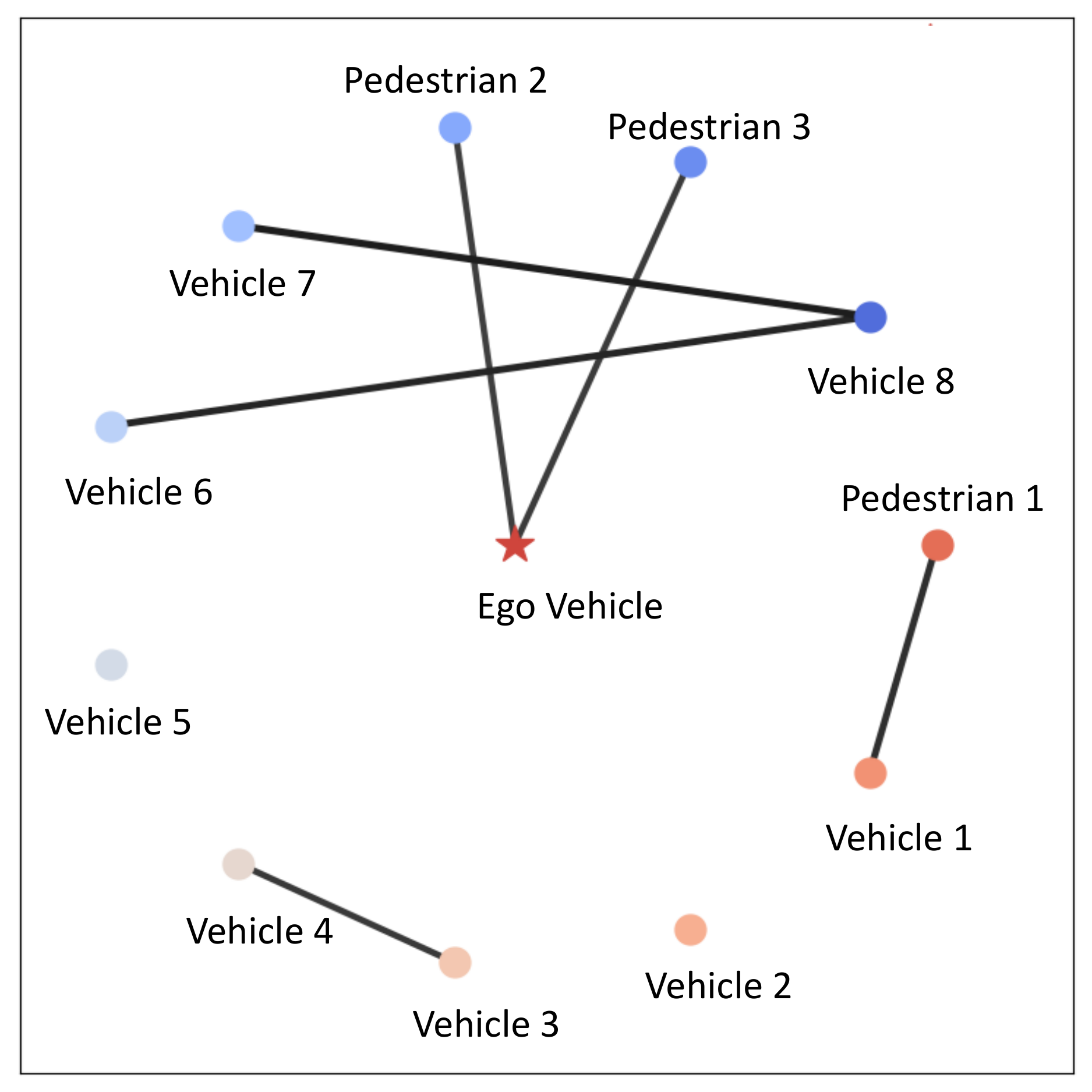}
\endminipage\hfill
\minipage{0.64\columnwidth}
 \vspace{0pt}
  \includegraphics[height=0.55\linewidth,page=1]{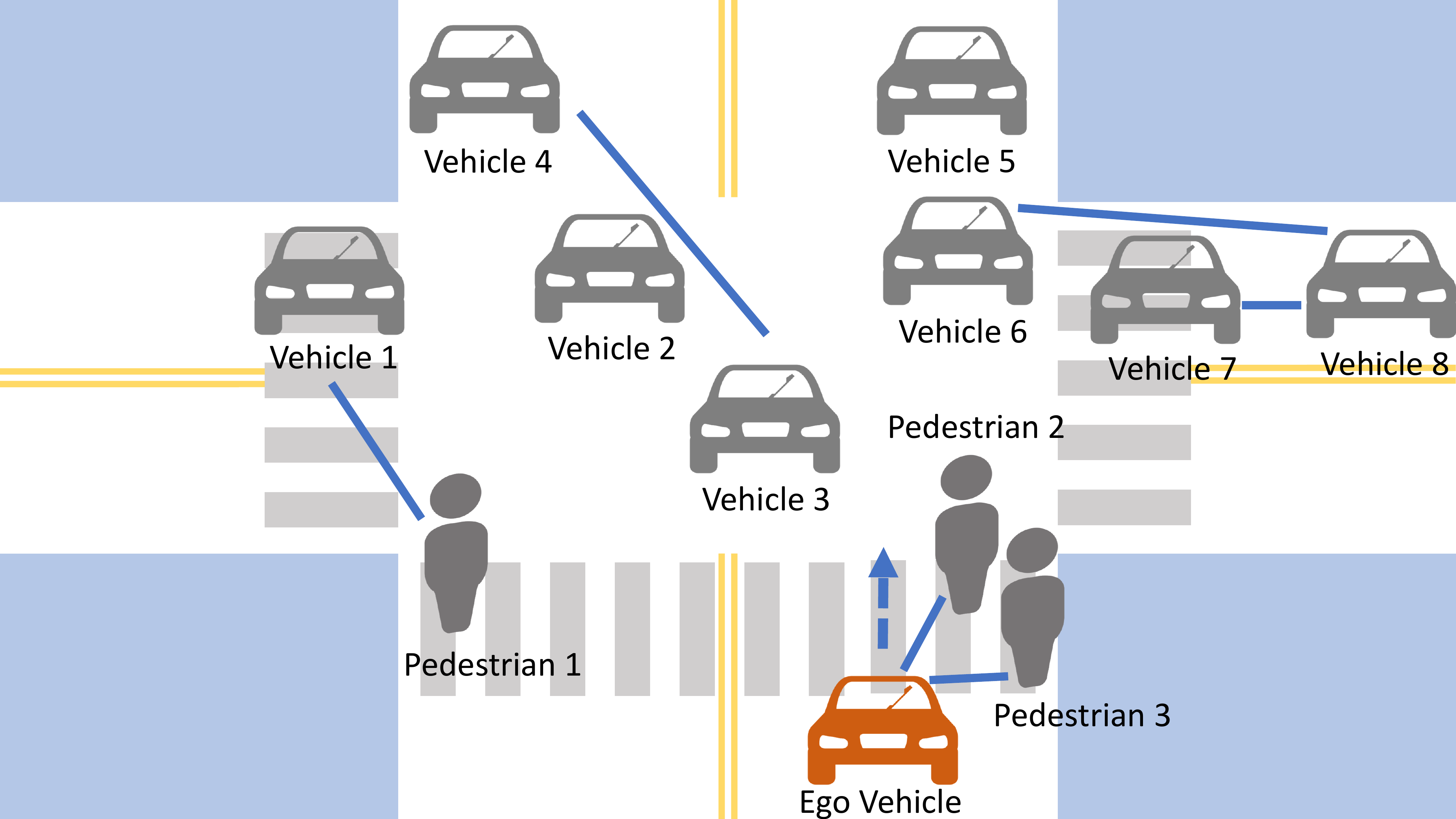}
\endminipage
}
\subfloat[Learned affinity matrix and BEV visualization.\label{fig:visualization_scene_representation_d}]{
\minipage{0.36\columnwidth}
  \includegraphics[width=\linewidth,page=2]{figures/bev_graph/graph_vis_graph.pdf}
\endminipage\hfill
\minipage{0.64\columnwidth}
 \vspace{0pt}
  \includegraphics[height=0.55\linewidth,page=2]{figures/bev_graph/graph_vis_configration.pdf}
\endminipage
}
\centering
 \caption{Visualization of Learned \textit{Ego-Thing Graph}. In (a) and (b), all detected objects are shown in bounding boxes with different colors. Every object is also associated with its class name and index. The left figures in (c) and (d) are the learned affinity matrices visualized in a graph structure. Each circle in the graph corresponds to a thing object in the frame and the driver is represented by a star. The edge linking two nodes represents the interactive relation among those nodes. A bird's-eye-view representation is shown on the right, providing more details of the traffic configurations.}
 \label{fig:visualization_scene_representation}
\end{figure*}


\noindent\textbf{Visualization of Learned Ego-Thing Graph.} 
We visualize learned relations in \textit{Ego-Thing Graph}.  
The left figures in Fig.~\ref{fig:visualization_scene_representation_c} and~\ref{fig:visualization_scene_representation_d} show two \textit{Ego-Thing Graphs} in two reactive scenarios.
We discard false positive detections. 
Each circle in the left-hand side figure corresponds to a \textit{Thing} object in the frame and the driver is represented by a star.
Notice that in the learned affinity matrix, every entry ${G}^{ET}(i,j)$ is a scalar ranging from 0 to 1, representing how object $j$ affects object $i$.
We take the average of ${G}^{ET}(i,j)$ and ${G}^{ET}(j,i)$ to represent the interactive relations between object $i$ and $j$ and draw an undirected edge between two nodes if the average value is larger than 0.2.
We manually create corresponding bird's-eye-view (BEV) maps in Fig.~\ref{fig:visualization_scene_representation_c} and~\ref{fig:visualization_scene_representation_d} to better represent the traffic configuration.
Fig.~\ref{fig:visualization_scene_representation_a} is an example of when the driver stops at a four-way intersection waiting for the traffic light.
The \textit{Ego-Thing Graph} captures interesting relations.
As the driver intends to pass through the intersection, two pedestrians ahead will influence driver behavior.
Vehicle 6, Vehicle 7 and Vehicle 8 in the distance are likely to interact with each other as they are in the same traffic lane.
We also observe false positives such as the detected interactions between Pedestrian 1 with Vehicle 1, and Vehicle 3 with Vehicle 4.
Explicit motion modeling of traffic participants could be used to mitigate these errors. 

Fig.~\ref{fig:visualization_scene_representation_b} depicts that the driver intends to turn left while yielding to an oncoming vehicle.
Our model detects interactions between the driver and three objects on the left side {}---{} Pedestrian 1, Pedestrian 2 and Vehicle 1.
For Pedestrian 3, Pedestrian 4 and Vehicle 2 on the right, although they do not interact with the driver, our model reveals that they influence each other because they are adjacent.

\begin{figure*}[t!]
\subfloat[Crossing Vehicle\label{fig:risk_score_a}]{
\minipage{0.54\columnwidth}
  \includegraphics[width=\linewidth,trim = {0 0 0 0},clip]{figures/riskscore/201706061647_010_cauese_crossing_vehicle_14505_14615_14561_image.png}
\endminipage
\minipage{0.46\columnwidth}
  \includegraphics[width=\linewidth,trim = {18 18 18 18},clip]{figures/riskscore/201706061647_010_cauese_crossing_vehicle_14505_14615_14561_score.png}
\endminipage}
\subfloat[Crossing Pedestrian\label{fig:risk_score_b}]{
\minipage{0.54\columnwidth}
  \includegraphics[width=\linewidth,trim = {0 0 0 0},clip]{figures/riskscore/201706061536_058_cauese_crossing_pedestrian_80511_81327_80957_image.png}
\endminipage
\minipage{0.46\columnwidth}%
  \includegraphics[width=\linewidth,trim = {18 18 18 18},clip]{figures/riskscore/201706061536_058_cauese_crossing_pedestrian_80511_81327_80957_score.png}
\endminipage 
}
\hfill
\subfloat[Parked Vehicle\label{fig:risk_score_c}]{
\minipage{0.54\columnwidth}
  \includegraphics[width=\linewidth,trim = {0 0 0 0},clip]{figures/riskscore/201703081008_050_cauese_parked_vehicle_55930_56144_55986_image.png}
\endminipage
\minipage{0.46\columnwidth}
  \includegraphics[width=\linewidth,trim = {18 18 18 18},clip]{figures/riskscore/201703081008_050_cauese_parked_vehicle_55930_56144_55986_score.png}
\endminipage}
\subfloat[Congestion \label{fig:risk_score_d}]{
\minipage{0.54\columnwidth}
  \includegraphics[width=\linewidth,trim = {0 0 0 0},clip]{figures/riskscore/201706061140_049_cauese_congestion_109891_112577_109947_image.png}
\endminipage
\minipage{0.46\columnwidth}
  \includegraphics[width=\linewidth,trim = {18 18 18 18},clip]{figures/riskscore/201706061140_049_cauese_congestion_109891_112577_109947_score.png}
\endminipage}

\caption{
%
Risk Assessment Visualization. 
On the left, all detected objects are shown in bounding boxes with different colors. The risk score of each object is depicted in a bar chart on the right. The color of each bar is one-to-one matched to the bounding box. We use a black horizontal line to indicate the predicted \textit{Go} score of the driver without applying any intervention.
}
\label{fig:risk_score}
\end{figure*}

\noindent\textbf{Risk Assessment.}
Our framework is able to perform risk assessment.
%
We visualize objects' risk scores in Fig.~\ref{fig:risk_score} under different reactive scenarios.
All detected objects are encased in bounding boxes with different colors.
Their risk scores are shown in a bar chart with color.
The risk score of an object is equivalent to the predicted confidence score of \textit{Go} after the object is removed.
A higher score of \textit{Go} means that the object has a higher chance of influencing driver behavior.
We use a black horizontal line to indicate the predicted confidence score of \textit{Go} when the input is not intervened.  
%
In all these cases, confidence scores are smaller than 0.5, representing correct driver response prediction by the proposed model.
Favorable risk assessment results are demonstrated.